\documentclass[10pt, a4paper, twocolumn]{article}

\usepackage{cite}
\usepackage{amsmath,amssymb,amsfonts}
\usepackage{algpseudocode}
\usepackage{graphicx}
\usepackage{svg}
\usepackage{booktabs}
\usepackage{multirow}
\usepackage{textcomp}
\usepackage{xcolor}
\usepackage{comment}
\usepackage{soul}

\DeclareMathOperator*{\argmin}{min}
\usepackage{geometry}
\date{20-11-2023}

\title{\textbf{The Safety Shell: An Architecture to Handle Functional Insufficiencies in Automated Driving}}
\author{
    C.A.J. Hanselaar$^{1}$\thanks{$^{1}$Department of Mechanical Engineering, Control Systems Technology Section, Eindhoven University of Technology, P.O.~Box 513, 5600 MB Eindhoven, The Netherlands (c.a.j.hanselaar@tue.nl)}, 
    E. Silvas$^{1,2}$\thanks{$^{2}$TNO, Dutch Organization for Applied Scientific Research}, 
    A. Terechko$^{3}$\thanks{$^{3}$NXP Semiconductors N.V}
    and W.P.M.H. Heemels$^{1}$
   }
    
\begin{document}
\maketitle
\begin{abstract}
\noindent {To enable highly automated vehicles where the driver is no longer a safety backup, the vehicle must deal with various Functional Insufficiencies (FIs). 
Thus-far, there is no widely accepted functional architecture that maximizes the availability of autonomy and ensures safety in complex vehicle operational design domains. 
In this paper, we present a survey of existing methods that strive to prevent or handle FIs.
We observe that current design-time methods of preventing FIs lack completeness guarantees. 
Complementary solutions for on-line handling cannot suitably increase safety without seriously impacting availability of journey continuing autonomous functionality.
To fill this gap, we propose the Safety Shell, a scalable multi-channel architecture and arbitration design, built upon preexisting functional safety redundant channel architectures.
We compare this novel approach to existing architectures using numerical case studies.
The results show that the Safety Shell architecture allows the automated vehicle to be as safe or safer compared to alternatives, while simultaneously improving availability of vehicle autonomy{, thereby increasing the possible coverage of on-line functional insufficiency handling.}}
\end{abstract}
\vspace{5pt}

\section{Introduction}
\label{section:Introduction}
\begin{figure*}[ht!]
    \centering
    \includegraphics[width=\textwidth]{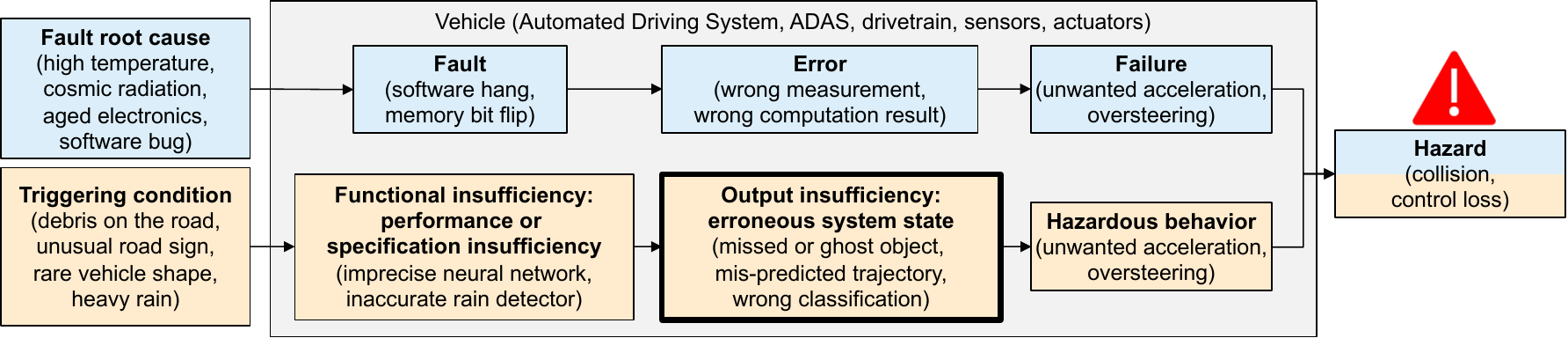}
    \caption{The distinction between Faults and Functional Insufficiencies, reused from \cite{Fu2023} with permission.}
    \label{fig:FIvsFaultsFromTerechko}
\end{figure*}

The promise of improved safety, sustainability and comfort has driven an ongoing desire for highly automated vehicles (vehicles fitted with at least one Automated Driving System, or ADS). 
Automated vehicles can help reduce traffic casualties caused by human mistakes, a significant worldwide problem~\cite{WHO2018}.
Next, automated vehicles can accelerate the implementation of sustainable transportation methods such as mobility-as-a-service concepts \cite{Hogeveen2021}.
However, several challenges still remain before highly automated driving vehicles can become widely available.
Most importantly, automated vehicles need to become safe enough to allow a transfer of responsibility for safety from the driver (SAE L2 driving or below) to the vehicle itself (SAE L3 or above)~\cite{SAEJ3016_2021}. 
Since ensuring that a vehicle never has an accident is only possible by not participating in traffic \cite{Shalev-Shwartz2017}, the need for the availability of journey continuing automated driving functions (often abbreviated to availability) will lead to some non-zero chance of collisions with automated vehicles.
To be accepted by society, an automated vehicle will have to be at least safer than a responsible human driver~\cite{koopman2022continuous}, yet provide similar levels of comfort.
To get an understanding of the current limits of automated vehicle capabilities with regards to safety, causes of supervised automated driving test disengagements are studied~\cite{DMV2021}.
Of the investigated disengagements, only $14\%$ was attributed to the occurrence of a fault, while $69\%$ was attributed to the occurrence of so-called Functional Insufficiency (FI) in the ADS~\cite{Fu2023}.
A FI is \textit{an inherent limitation in a function inside the studied system that leads to hazardous (dangerous) system behaviour}, exposed by the specific circumstances that the system is operating in, as defined in the Safety Of The Intended Functionality (SOTIF) standard~\cite{iso21448}.
To clarify the distinction and similarities between systematic faults covered in ISO26262~\cite{iso26262} and FIs, we refer to Fig. \ref{fig:FIvsFaultsFromTerechko}. 
To handle FIs, SOTIF focuses on design-time eradication of FIs by identifying them through iterative testing and redevelopment, thereby reducing the number of unknown FIs that remain in the system. 
Given that FIs currently represent the dominant cause for automated driving test disengagements, knowledge of FI handling methods is required to develop a safe ADS.
However, we are not aware of a sufficiently comprehensive survey on this problem.

Therefore, we identify the gap of current approaches that aim to handle FIs during either development-time or run-time through a survey and we propose a novel architecture to handle remaining FIs during run-time with minimal impact on system availability.
Our contributions can be summarized as follows: 
\begin{itemize}
    \item[-] We analyze and identify the gap in FI handling approaches through a literature survey.
    \item[-] We introduction a novel multi-channel cross-checking architecture for which we coin the term Safety Shell.
    \item[-] We introduce a method to perform a relative safety and availability evaluation of different architectures.
    \item[-] We assess the Safety Shell in comparison to other architectures using extensive numerical simulations.
\end{itemize}

The remainder of this paper is organised as follows. Section \ref{section:stateoftheart} will provide an overview of the current state of art regarding FI prevention, detection and mitigation and {identify the} remaining gaps.
To bridge these gaps, 
Section \ref{section:SafetyShellDesignPattern} introduces the Safety Shell.
The numerical case-studies used to assess different architecture performances are shown in Section \ref{section:NumericalSims}.
Finally Section \ref{section:discussion} provides a discussion of the results, while Section \ref{section:conclusion} summarizes our findings.

\section{State of the art}
\label{section:stateoftheart}
\begin{figure}
    \centering
    \includegraphics[width=7.5cm]{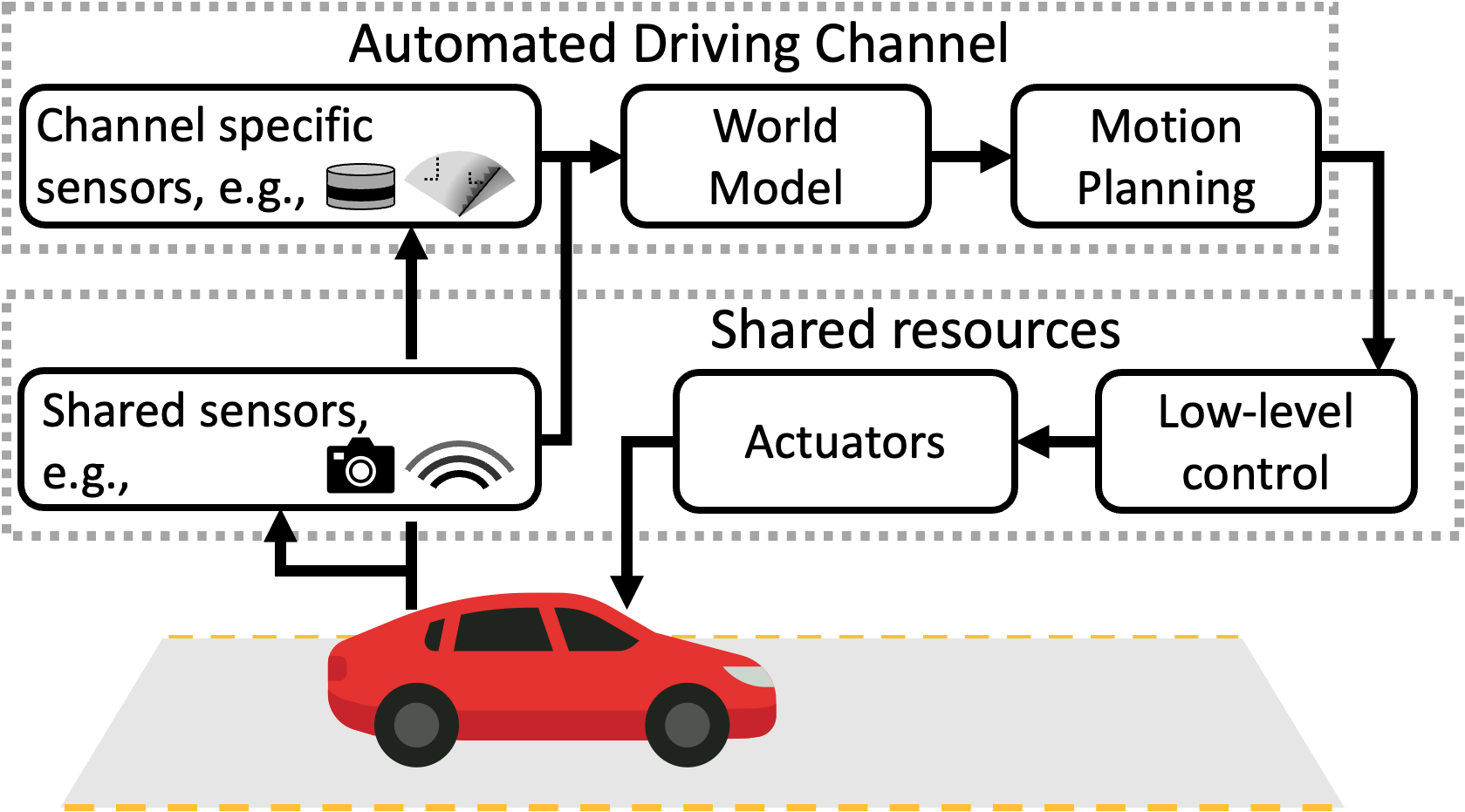}
    \vspace{-7 pt}
    \caption{A simplified automated vehicle system architecture. Sensors (shared and channel specific) feed information through the World Model to the Motion Planning functions to create trajectory plans. These provide actuation setpoints, as executed by low-level and actuators.}
    \label{fig:general_architecture_overview}
    \vspace{-5 pt}
\end{figure}

To facilitate a common understanding, we introduce a very high level architecture of an ADS in a highly automated vehicle in Fig \ref{fig:general_architecture_overview}.
Following e.g. \cite{Furst2018,Ishigooka2018,Fu2020}, multiple redundant ADSs can be included in an automated vehicle. 
We refer to these as Automated Driving (AD) channels.
We therefore distinguish the components that are shared between different redundant systems e.g., shared sensors, low-level controllers and actuators, and AD channel specific components, e.g. channel specific sensors and functions. 
An AD channel capable of L2 or higher automation at least includes a {World Model} (WM) function, comprised of e.g., various multi-sensor fusion subfunctions for ego localisation and state estimation, object detection, identification and  motion prediction and road modelling. 
The WM output is passed to the {Motion Planning} (MP) functions, 
which are responsible for the coordination of vehicle decision making and motion (e.g., route, path and trajectory planners and motion controllers).
The MP provides setpoints to vehicle low-level control and actuators (e.g., steering angle setpoint, acceleration setpoint).

\subsection{FI handling during development}
\label{section:FiHandlingDuringDevelopment}
To follow and comply with SOTIF, most FI handling for autonomous vehicle development is focused on first identifying triggering conditions, then improving functions to better handle the triggering condition, and finally verifying and validating the improved functionality.
For example, if initial testing identifies that a WM in an ADS undersizes the bounding boxes used to allocate occupied space to identified pedestrians, the responsible WM function is be redeveloped~\cite{cheng2022logically,schuster2022formally,gerchinovitz2022object} and retested. 
Unfortunately \cite{gerchinovitz2022object} inadvertently shows a potential FI in the first figure of their study, where a partially occluded pedestrian is not identified by their updated system, indicating some undiscovered FIs remain in their WM, despite their improvements.

Classical approaches \cite{kirovskii2019driver,ferrari2022criteria} intend to provide improved guidelines for the iterative development process, to try to attain sufficient safety prior to release of a function.
Examples of these works are e.g., studies that use Decision Safety Analysis through scenario-based assessments, to evaluate if the system decision logic is sound~\cite{osborne2022analysing}, or studies to extend a framework for safety case patterns based on ISO 26262 to machine learned functions and possibly occurring FIs~\cite{wozniak2020safety, radlak2020organization, zeller2022component}. 
Similarly, design-time handling of potential FIs is performed via an architectural pattern for each function that assures safety~\cite{kochanthara2021functional}, in extension of the functional safety requirement handling in ISO 26262. 
However, an evaluation of the reported architectural patterns to assure safe functions shows that the proposed patterns (sanity check, barrier coding, heartbeat checking and condition monitoring)~\cite{kochanthara2021functional} cannot detect or mitigate FIs related to, e.g., object presence or state state or drivable space misrepresentations. 
Both this counterexample and the prior missed pedestrian counter example highlight the difficulty of ensuring completeness at design time, if an important hazard or cause is unknown during development, despite accepted state-of-art methods, field-leading expert cooperation, robust design elements and architectures.
Next, the large number of unique out-of-the-ordinary situations encountered in real traffic makes accumulating sufficient proof of safety through only design-time testing a near-impossible task~\cite{Koopman2020c}.
Each of these rare and unique situations contributes very little to overall risk, but, by virtue of the number of different unique situations, together they represent a significant portion of daily driving called the heavy tail~\cite{Koopman2020c,Koopman2022HowSafe}. 

Recently, approaches such as~\cite{korssen2017systematic,saberi2020beyond, Fu2020} highlight the benefit of formal verification of created functions, as they provide proofs that systems will behave as expected no matter the inputs to the functions.
Unfortunately, the inherent computational limitations to formal verification techniques and the difficulty of applying it to uncertain environment perceptions seem to prevent its application to full ADS verification.
Finally, we need to consider the fact that an automated vehicles is expected to operate for several years.
The driving scenarios relevant in 5 or 10 years are likely different from the scenarios the ADS is designed to cope with.
To fulfil an obligation to safety, ADS developers and automated vehicle manufacturers will need to monitor leading metrics (e.g., Safety Performance Indicators~\cite{ul4600,koopman2022continuous}) during the lifetime of the vehicle and, if the Safety Performance Indicators suggest it, adjust the ADS so it can continue to operate safely.

In summary, development-time exclusive approaches to eliminate FIs cannot guarantee completeness, nor are they inherently suited for quantitative safety or risk analysis that is required for SAE L3 and beyond~\cite{ul4600}.
Despite the efforts of design and development teams following SOTIF and functional safety standards, the infinite variations in scenarios will ensure that a significant number of untested scenarios will remain after an economically viable development period~\cite{Koopman2020c}.
Therefore, unknown FIs will persist after a design is finished, necessitating highly automated vehicles to be fitted with both some kind of online FI handling approach as well as a method to continuously assess if the automated vehicle is still safe enough to operate, through the tracking of leading safety metrics~\cite{koopman2022continuous,ul4600}.

\vspace{-5pt}
\subsection{Principles of online FI handling}
\label{section:heterogeneiety}
Detecting a FI requires knowledge of the ground truth. 
In supervised automated vehicle testing, this is approximated by a human test driver.
For automated vehicles without supervision at least two heterogeneous AD channels {and a method for cross-comparison and arbitration are required.}
{This necessitates a discussion on design patterns and architectures that enable such methods.}
A heterogeneous AD channel function can detect a difference in its sister function output, but because none directly measure the ground truth, it is unclear which (if not both) of the functions suffers from a FI. 
\textit{Heterogeneity of functions is therefore a necessary but insufficient requirement to detect a FI during run-time.}
Similarly, to allow for automated monitoring of the safety of an implemented AD channel after release, some kind of detection of disagreement will be required.
Acknowledging this limitation, the aim of systems that apply heterogeneous functions is to detect divergent behaviour.
This indicates that fault-detection patterns such as homogeneous 2-out-of-2 systems \cite{Armoush2010,oliveira2023architectures} are unable to detect FIs because they lack heterogeneity.

Heterogeneity in highly AD channel functions affects:
\begin{enumerate}
    \item The way an AD channel perceives the environment;
    \item The way an AD channel chooses to act based on the perception of its environment.
\end{enumerate}
These differences lead to both varying Operational Design Domain (ODD) limits and different unknown FIs that remain after development. 
\begin{figure}
    \centering
    \includegraphics[width=8.4cm]{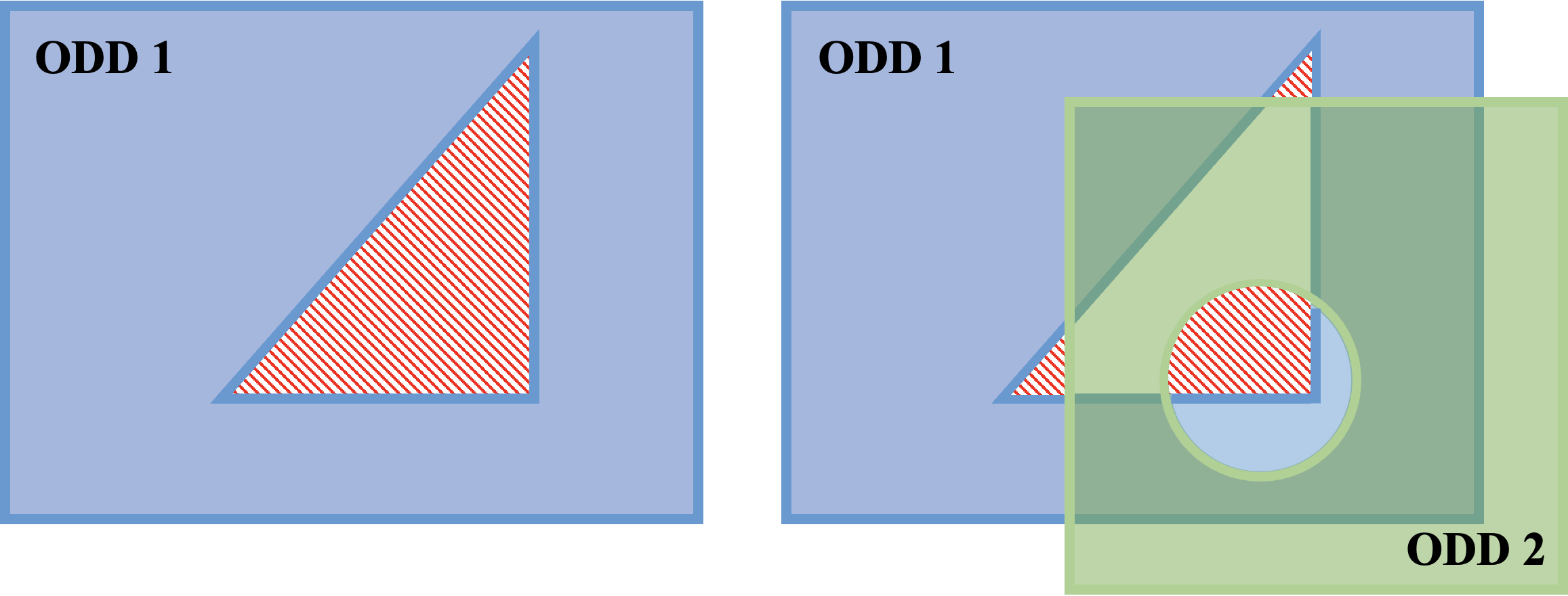}
    \vspace{-4 pt}
    \caption{Abstract representation of the heterogeneous ODDs of functions, represented by the shaded areas. The gaps in the shaded areas represent remaining unknown FIs, with the oblique lines indicating FIs that are undetectable during run-time.}
    \vspace{-5pt}
    \label{fig:WM_Comparisons}
\end{figure}  
In Fig. \ref{fig:WM_Comparisons} we show an abstract representation of an AD channel function and its ODD on the left, with the remaining unknown FIs indicated in the triangular area with oblique lines. 
The ODD of a function from a second AD channel is superimposed on the first on the right side of the figure and its remaining FIs are indicated with a circle. 
To reduce the number of remaining FIs and, consequently, increase the safety of the vehicle, a suitable cross-checking comparison and mitigation method is required.
The result of a successful method is indicated by the reduced area with oblique lines on the right. 
However, if the comparison and mitigation is too strict, the usable area of the combined functions can shrink to exclude areas of disagreement, reducing the overall availability of functions.

In summary, the presence of heterogeneous functions is a necessary but insufficient condition for FI detection. 
Through the application of a number of heterogeneous redundant channels and the application of cross-checking methods, the chance of possible FI detection can be increased, but this must be weighted against the impact on availability.

\vspace{-3pt}
\subsection{Implementations for online FI handling}
\label{section:existingPatterns}
All design patterns for FI handling try to reduce the number of unhandled possible FIs through testing for possible output insufficiencies or function differences~\cite{weast2020sensors,salay2022safety,Mehmed2019,Mehmed2020,Torngren2019,Furst2018,Fruehling2019} (see Fig. \ref{fig:FIvsFaultsFromTerechko}). 
As these tests occur during operation, these generalizable architectures can contribute to reductions in dangerous behaviour by the automated vehicle, even for the unknown FIs.

\begin{figure}[t]
    \centering
    \includegraphics[width=6cm]{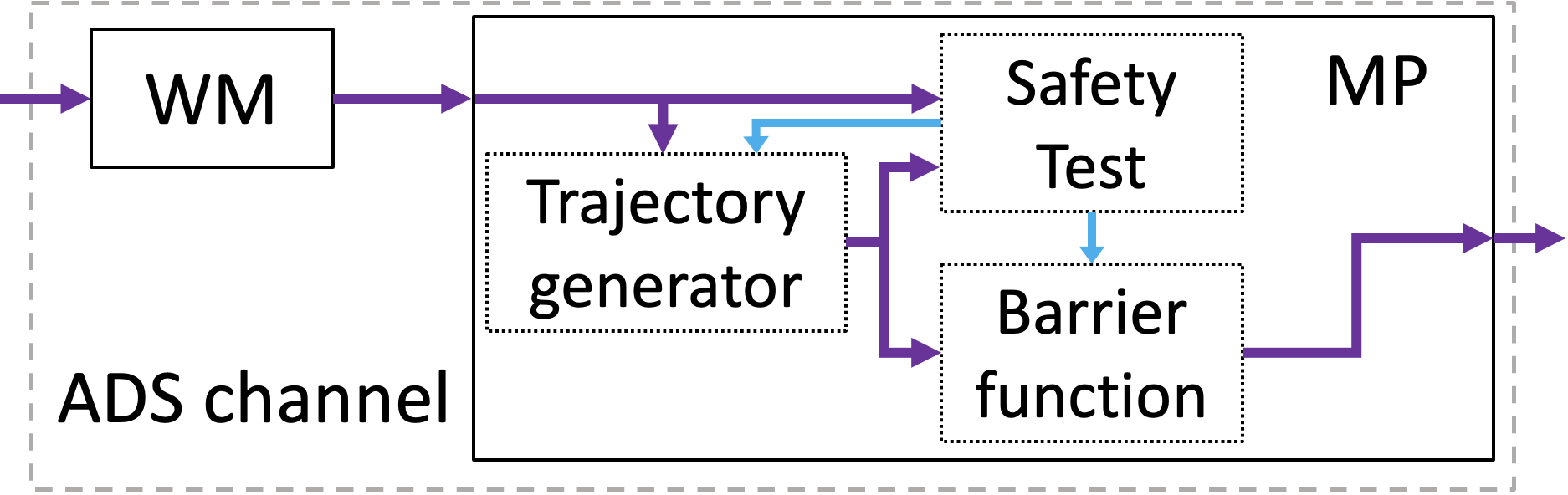}
    \caption{Representation of the single-channel Monitor-Actuator architecture, with the Safety Test feedback lines in blue.}
    \label{fig:SimpleBarrierArchitecture}
    \vspace{-8pt}
\end{figure}

The most limited approach of FI detection can be narrowed down to a \textit{Monitor-Actuator} design pattern~\cite{Armoush2010,Koopman2016,konighofer2020shield} (MA, also known as \textit{doer-checker} or \textit{shield} pattern), as shown in Fig. \ref{fig:SimpleBarrierArchitecture}, with the arrows indicating flow of information.
Note that the Sensors shown in Fig. \ref{fig:general_architecture_overview} are not shown in the following figures to simplify them.
The safety test uses the information of the nominal WM to assess the safety of the generated trajectory.
If it judges the trajectory as safe, it allows the barrier function to pass the trajectory {on}.
The safety test function has a feedback line to the trajectory generator, used to provide counter-examples that invalidate the safety of a proposed trajectory and indicate a required recalculation~\cite{Decastro2018,Wang2019}. 
Integrated guarantees for legally safe motion planning output~\cite{pek2020FailSafeTrajectory,pek2020fail,Jackson2019} have been proposed as a way to stay reasonably safe as well as to ensure no blame lies with automated vehicles in case of collisions.
For the purpose of our discussion, we refer to this pattern as a \textit{single-channel MA} pattern. 

The same single-channel MA pattern is used in the Responsibility Sensitive Safety (RSS) architecture~\cite{Shalev-Shwartz2017} and the similar Safety Force Field (SFF)\cite{Nister2019}.
Both prescribe longitudinal and lateral safety distances as a function of ego and other object states and presumed acceleration capabilities, and enforce mitigating actions (e.g., braking) if nominal MP output would violate those safe distances.
These approaches assure safety if all vehicles follow them~\cite{Shalev-Shwartz2017}, but this utopia is unlikely to ever by attained if humans interact with the automated vehicles.
Beyond this assertion, certain complex environments can require risk mitigation in ways that simpler rule-based algorithms such as RSS or SFF do not allow \cite{Fruehling2019}. 

Understandably, none of the works on safe motion planning can claim to ensure the safety of a vehicle, if perception is lacking, e.g., a certifiably safe trajectory is only safe with respect to observed objects and predicted object motions.
This is especially relevant for assuredly safe motion planning (e.g., \cite{pek2020FailSafeTrajectory,pek2020fail,Jackson2019}) or multi-hypothesis planners (e.g., \cite{pek2020FailSafeTrajectory,pek2020fail}), or even the simpler safety rule-based approaches~\cite{Shalev-Shwartz2017,Nister2019}, as very late object detections can result in the absence of a feasible safe trajectory or response. 
{Single WM approaches without a backup planner are therefore more suitable for low-speed vehicles in more controlled circumstances.}

\begin{figure}[t]
    \centering
    \includegraphics[width=8.5cm]{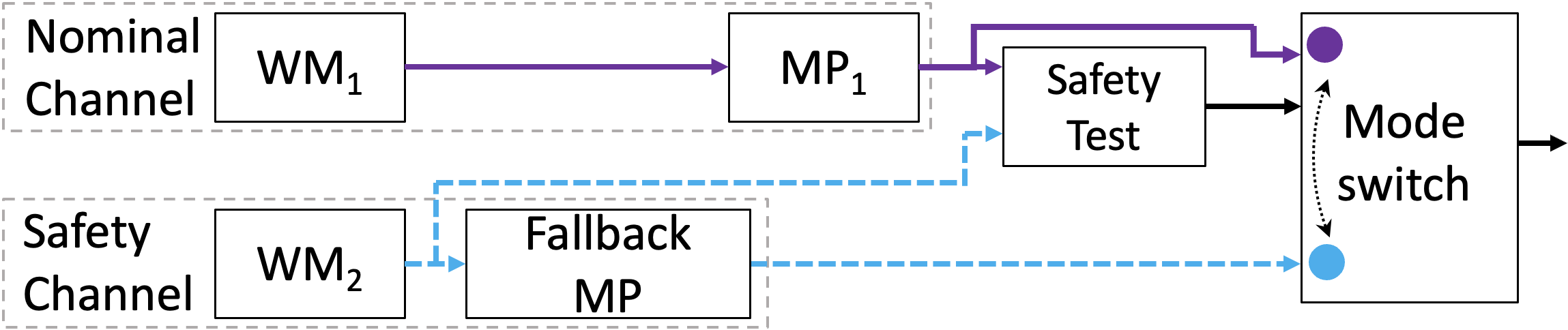}
    \caption{Representation of a Monitor-Actuator architecture with a nominal and a safety channel. The mode switch indicates a transition from mission-continuing ADS functionality to a safe-state attaining fallback function.}
    \label{fig:Mehmed2019}
\end{figure}

To also allow WM-caused FI handling {and retain a fallback MP in case of persistent MP issues}, the MA design pattern shown in Fig. \ref{fig:SimpleBarrierArchitecture} is augmented to Fig. \ref{fig:Mehmed2019}, based on~\cite{Mehmed2019}.
In Fig. \ref{fig:Mehmed2019} the signals from the safety channel are shown in blue with dashed lines, to distinguish them from the nominal channel signals.
This MA pattern and variations of it test for safety violations of the planned trajectory, not just by comparing to the nominal WM but also by comparing to a dedicated separate safety WM~\cite{Mehmed2019,Mehmed2020,Torngren2019,weast2020sensors,WaymoSafety2021}.
Such safety WMs are often simplified compared to nominal WMs, to obtain a higher belief in the correct operation of these functions \cite{Mehmed2019}.
Going forward, we will be referring to this pattern simply as the \textit{MA} pattern. 
If a safety threshold violation occurs, a safety fallback motion planner is activated~\cite{Mehmed2019,Mehmed2020,Torngren2019,WaymoSafety2021}, or output is restricted to an assumed safe setpoint envelope~\cite{weast2020sensors}.
Often these safety fallback motion planners are kept simple to ensure that their behaviour is predictable, such as automatic emergency braking (AEB) systems \cite{WaymoSafety2021}.
We refer to this fallback strategy as an emergency response, and to the trajectory that governs this as the escape trajectory.

\begin{table*}[]
\centering
\caption{Summary of identified FI-handling architectures, with their advantages and disadvantages}
\label{table:overviewOfArchitectures}
\resizebox{\textwidth}{!}{%
\begin{tabular}{@{}llllll@{}}
\toprule
\textbf{Design pattern} & \textbf{Detection method of FIs} & \textbf{Interventions} & \textbf{Advantages} & \textbf{Disadvantages} & \textbf{Associated publications} \\ \midrule
\multicolumn{1}{|l|}{\begin{tabular}[c]{@{}l@{}}Single-channel \\ MA pattern\\ Fig. \ref{fig:SimpleBarrierArchitecture}\end{tabular}} & \multicolumn{1}{l|}{\begin{tabular}[c]{@{}l@{}}Safety test w.r.t nominal \\ WM or formal guarantees\\ of nominal MP\end{tabular}} & \multicolumn{1}{l|}{\begin{tabular}[c]{@{}l@{}}Not publishing\\ trajectory / \\ restricted output \\ {control setpoints} \end{tabular}} & \multicolumn{1}{l|}{\begin{tabular}[c]{@{}l@{}}Simple design,\\ robust against \\ MP FIs\end{tabular}} & \multicolumn{1}{l|}{\begin{tabular}[c]{@{}l@{}}WM FIs undetectable, in case\\ of not publishing {then there is}\\ no clear fallback available\end{tabular}} & \multicolumn{1}{l|}{\cite{Decastro2018, Wang2019,pek2020FailSafeTrajectory, pek2020fail, Jackson2019,Shalev-Shwartz2017, Nister2019}} \\ \midrule
\multicolumn{1}{|l|}{\begin{tabular}[c]{@{}l@{}}MA pattern,\\ one safety WM\\ Fig. \ref{fig:Mehmed2019}\end{tabular}} & \multicolumn{1}{l|}{\begin{tabular}[c]{@{}l@{}}Safety test of nominal \\ MP w.r.t. safety WM\end{tabular}} & \multicolumn{1}{l|}{\begin{tabular}[c]{@{}l@{}}Switch to \\ fallback MP\end{tabular}} & \multicolumn{1}{l|}{\begin{tabular}[c]{@{}l@{}}Can detect FIs caused \\ by nominal WM and MP \\ if not simultaneously \\ occurring in safety WM\end{tabular}} & \multicolumn{1}{l|}{\begin{tabular}[c]{@{}l@{}}Sensitive to FP safety test fails,\\ limited fallback capability, \\ limited WM FI coverage through\\ single safety WM addition\end{tabular}} & \multicolumn{1}{l|}{\cite{Mehmed2019}} \\ \midrule
\multicolumn{1}{|l|}{\begin{tabular}[c]{@{}l@{}}MA pattern with \\ safety WM and \\ WM comparison\\ Fig. \ref{fig:Torngren2018}\end{tabular}} & \multicolumn{1}{l|}{\begin{tabular}[c]{@{}l@{}}Safety test of nominal\\ MP w.r.t safety WM and\\ safety vs nominal WM\\ comparison\end{tabular}} & \multicolumn{1}{l|}{\begin{tabular}[c]{@{}l@{}}Switch to \\ fallback MP\end{tabular}} & \multicolumn{1}{l|}{\begin{tabular}[c]{@{}l@{}}Extra sensitive to leading \\ FIs (issues in WMs that \\ do {not} cause immediate \\ dangers), can detect FIs by\\ nominal WM and MP if\\ not simultaneously \\ occurring in safety WM\end{tabular}} & \multicolumn{1}{l|}{\begin{tabular}[c]{@{}l@{}}Highly sensitive to channel \\ differences, sensitive to FP safety\\ test fails, limited fallback \\ capability, limited WM FI\\ coverage through single \\ safety WM addition\end{tabular}} & \multicolumn{1}{l|}{\cite{Torngren2019}} \\ \midrule
\multicolumn{1}{|l|}{\begin{tabular}[c]{@{}l@{}}FWM pattern, \\ safety WM and\\ free space fusion\\ Fig. \ref{fig:Mehmed2020}\end{tabular}} & \multicolumn{1}{l|}{\begin{tabular}[c]{@{}l@{}}Safety test of nominal\\ MP w.r.t. safety WM\end{tabular}} & \multicolumn{1}{l|}{\begin{tabular}[c]{@{}l@{}}Switch to \\ fallback MP\end{tabular}} & \multicolumn{1}{l|}{\begin{tabular}[c]{@{}l@{}}Reduced sensitivity to FP \\ FIs in safety WM, can \\ detect FIs caused by \\ nominal WM and MP \\ if not simultaneously \\ occurring in safety WM\end{tabular}} & \multicolumn{1}{l|}{\begin{tabular}[c]{@{}l@{}}Limited WM FI coverage through\\ single safety WM, risk of \\ intertwined development of safety\\ and nominal WM, limited fallback\\ capability\end{tabular}} & \multicolumn{1}{l|}{\cite{Mehmed2020}} \\ \midrule
\multicolumn{1}{|l|}{\begin{tabular}[c]{@{}l@{}}RSS pattern, \\two safety WMs\\ Fig. \ref{fig:weast2020}\end{tabular}} & \multicolumn{1}{l|}{\begin{tabular}[c]{@{}l@{}}Safety test of nominal\\ MP w.r.t. both safety\\ WMs and nominal WM\end{tabular}} & \multicolumn{1}{l|}{\begin{tabular}[c]{@{}l@{}}Restricted output\\ control setpoints\end{tabular}} & \multicolumn{1}{l|}{\begin{tabular}[c]{@{}l@{}}Improved detection of FIs \\ caused by nominal WM \\ and MP if not simultaneously \\ occurring in any of the safety \\ WMs\end{tabular}} & \multicolumn{1}{l|}{\begin{tabular}[c]{@{}l@{}}Limited fallback capability, \\ pattern design limited to 2 \\ safety WMs\end{tabular}} & \multicolumn{1}{l|}{\cite{weast2020sensors,MobilEye2022TrueRedundancy}} \\ \bottomrule
\end{tabular}%
}
\end{table*}
\begin{figure}[t]
    \centering
    \includegraphics[width=8.5cm]{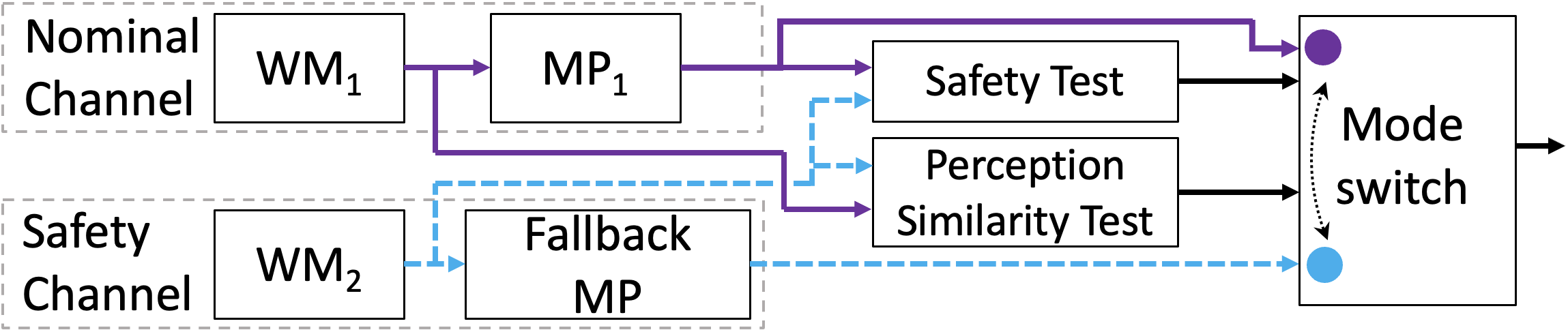}
    \caption{Representation of 2-channel Monitor-Actuator architecture by \cite{Torngren2019}, with a nominal channel, a safety channel and a perception similarity test.}
    \label{fig:Torngren2018}
\end{figure}  
\begin{figure}[t]
    \centering
    \includegraphics[width=8.5cm]{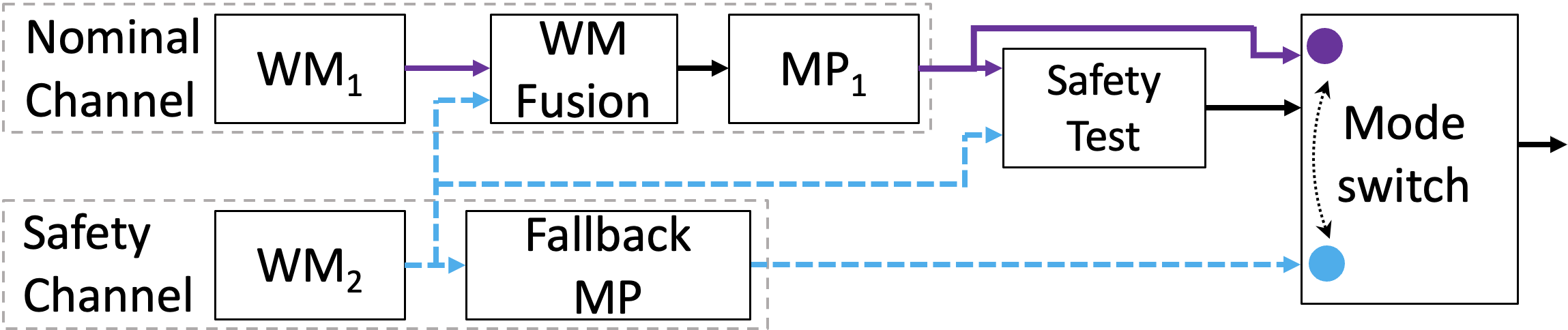}
    \caption{Representation of Fused World Model architecture by \cite{Mehmed2020}, with the coupled nominal and safety perception systems.}
    \label{fig:Mehmed2020}
\end{figure} 
\begin{figure}[ht!]
    \centering
    \includegraphics[width=7cm]{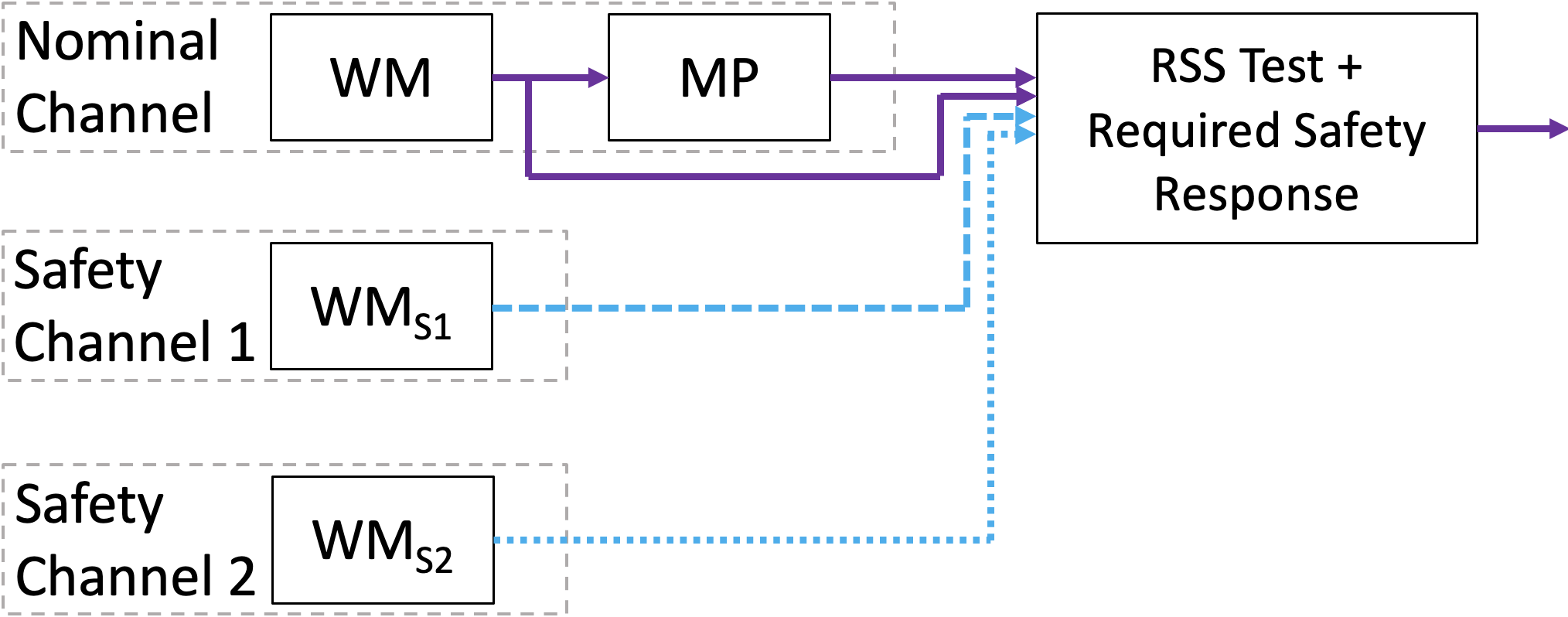}
    \caption{Representation of the proposed 3-World-Model approach by \cite{weast2020sensors,MobilEye2022TrueRedundancy}, with the single nominal channel and 2 safety WMs with their blue dashed and dotted signal lines.}
    \label{fig:weast2020}
    \vspace{-3pt}
\end{figure}  

These architectures and response strategies will impact the availability of nominal functionality.
In the case that the safety channel correctly identifies a hazardous planned trajectory and causes a mode switch to the safe-state attaining escape trajectory, the availability penalty is justified, though still undesired.
However, in case of a False-Positive (FP) issue (e.g., a ghost object detection by the safety WM), the availability penalty is not justified and the safety of the automated vehicle is decreased due to the harsh fallback MPs.
{The presence of this fallback allows for higher speeds to be planned and relaxes the requirement for nominal trajectories to end in a safe state to assure safety if the nominal MP or WM encounters an FI.
The system still requires the safe-state attaining fallback to be an acceptable solution in its entire ODD, meaning, e.g., highway driving is difficult with simple fallback planners, as stopping in the fast lane is not a safe state.}

Some design patterns go beyond testing nominal trajectories for unsafe motions and also test for environment perception disagreement~\cite{Torngren2019}, as shown in Fig. \ref{fig:Torngren2018}.
{This is similar to majority-voting approaches common in aviation, e.g., the triple-redundancy majority voting pattern} \cite{Armoush2010}. 
The main downside of this approach is the reduced availability of the nominal automated driving function due to the many non-significant differences between heterogeneous perception functions that can trigger the mode switch to the fallback channel and stop the vehicle. 
To overcome this~\cite{Mehmed2020} proposes to use the intersection of the free space from safety perception and nominal perception for all motion planning, as shown in Fig. \ref{fig:Mehmed2020}. 
This largely eliminates ghost-object-triggered mode switches to the fallback MP, as the nominal MP will have to take any FP obstacle detections by the safety WM into account in its motion planning.
Consequently, this architecture reduces disengagements at the cost of more conservative nominal MP behaviour.
We refer to this architecture as the\textit{ Fused World Model (FWM) }architecture.
As this architecture only includes a single simplified safety WM and fallback MP~\cite{Mehmed2019,Mehmed2020}, it is limited in its ability to increase the chance of detection of different FIs through the cross-comparisons to more heterogeneous WMs.
{The FWM solution is more robust against WM insufficiencies such as missed objects and, consequently, is able to plan safe motions despite a single occurring WM FI. 
It is therefore likely to be safer than the MA pattern, as its nominal MP is better able to find a safe path compared to fallback MPs.
Depending on the trust in the nominal MP, this approach may be suitable for a controlled ODD such as a limited speed highway automation. }

To allow increased probability of danger detection and increased redundancy in case of fault, a 3-channel architecture is proposed by \cite{weast2020sensors,MobilEye2022TrueRedundancy}, who suggest to have 2 fully heterogeneous safety perception channels in parallel with a nominal perception system, acting in a MA safety-threshold pattern shown in Fig. \ref{fig:weast2020}.
In this design both heterogeneous safety channels will be used to assess if the proposed actuation setpoints derived from the nominal MP respect the RSS limits.
Any violation of the RSS limits will ensure that the actuation setpoints are replaced by the required RSS safe action, e.g. brake with specified deceleration or perform limited lateral manoeuvres. 
In \cite{weast2020sensors} the authors mention that each individual safety perception channel could operate as a fully functional redundant channel in case of fault. 
However, based on \cite{MobilEye2022TrueRedundancy}, it appears that this entails rerouting the WM data from one of the safety channels into the same MP, instead of a parallel (possibly heterogeneous) MP.
Therefore for the purpose of this study we treat this architecture as a MA pattern with 2 safety WM systems.
Thanks to its integration with the RSS safety-indicator and envelope restriction, we refer to architecture as the \textit{RSS} architecture.
{Based on the single MP function, there is limited flexibility in case of a MP FI in the RSS architecture, yet the 3 heterogeneous WMs are expected to increase the chance of FI detection.
Consequently, the system may be suitable for limited-speed situations of more diverse natures than the FWM architecture, e.g., low-speed city manoeuvring, limited speed automation on more diverse roads, where suddenly stopping vehicles are not considered too obtrusive.}

\subsection{The gap in the state of the art}
\label{section:gap}
All designs covered in Section \ref{section:existingPatterns} are able to reduce some of the unknown FIs {that remain} in an ADS {after initial development}.
These designs, summarized in Table \ref{table:overviewOfArchitectures}, rely on a single nominal MP and a mode switch to a simple fallback MP or prescribed trajectory output restriction, triggered if the nominal system is judged as unsafe.
However, complex scenarios that cause nominal systems to encounter a FI, may be beyond the capability of simple fallback planners or RSS safe actions~\cite{Fruehling2019}.
For instance, stopping a highly automated vehicle in-lane on a highway is generally not a sufficiently safe state, nor is stopping on a railway intersection.
Next, by adding multiple safety WMs that can only trigger a mode switch to the fallback approach, the chance of a false positive trigger and consequently of unnecessary availability interruption increases, e.g., due to a false positive detection in one of the safety WMs.
Finally, of all architectures shown in Fig. \ref{fig:SimpleBarrierArchitecture}-\ref{fig:weast2020}, none are capable of continuing if a FI or fault occurs in the nominal motion planner, as all have to resort to either an envelope restriction or simple fallback MP.

To fill this gap we have distilled the following problem formulation:
How to develop an architecture capable of combining the best of several Automated Driving Channels in an automated vehicle, thereby increasing the safety of the automated vehicle without impeding the availability of automated driving functionality, when compared to alternative architectures?

\section{Safety Shell}
\label{section:SafetyShellDesignPattern}
To address the problem identified at the end of the previous section, we propose the Safety Shell in this paper, of which a preliminary conference version was presented in \cite{Hanselaar2022}.
Compared to~\cite{Hanselaar2022}, we provide, next to the survey in Section \ref{section:stateoftheart}, a more complete explanation of the Safety Shell concept, improve the arbitration algorithm, introduce the used risk calculations, clarify the process to determine key tuning parameters, introduce significantly more extensive numerical simulations, which includes both a three-channel implementation of the Safety Shell and comparisons to another proposed three-channel architecture, which were all not available in \cite{Hanselaar2022}.
\begin{figure}[ht]
    \centering
    \includegraphics[width=8.5cm]{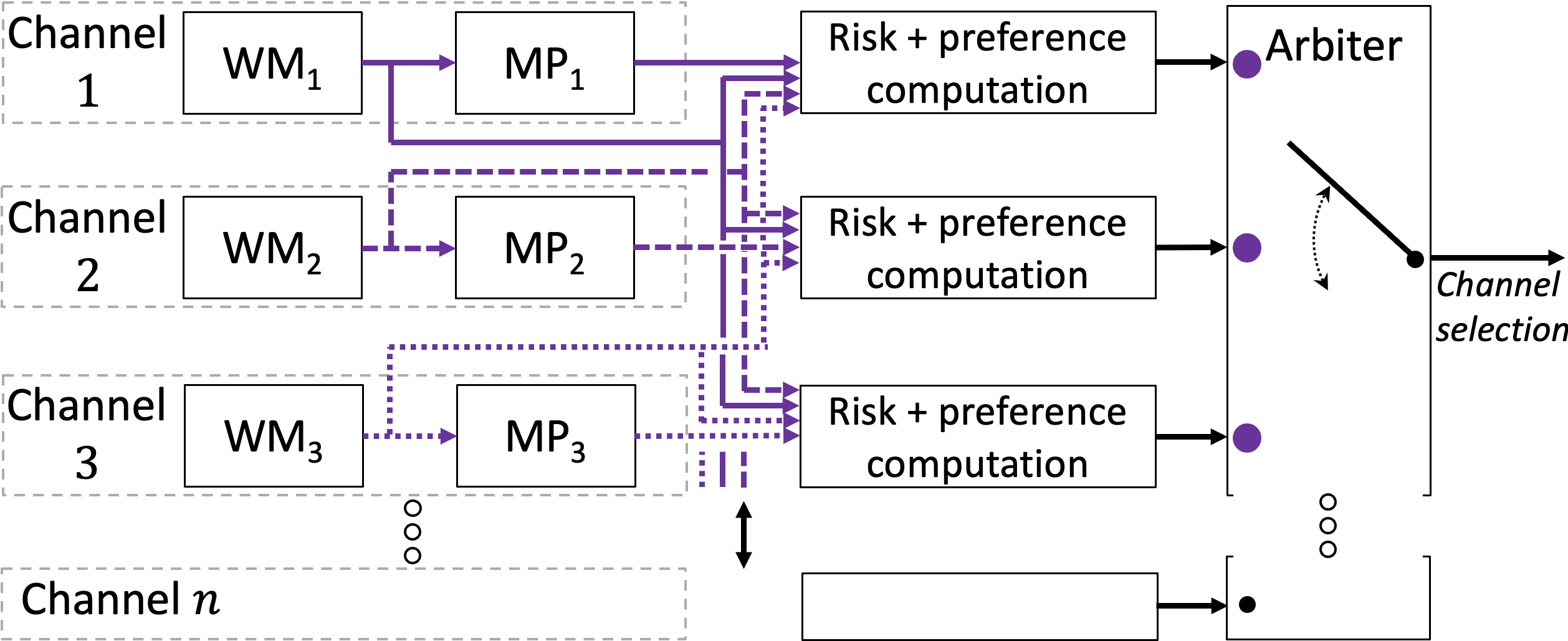}
    \caption{Representation of multi-channel Safety Shell architecture. Each channel is an ADS that is able to continue the journey of the vehicle. Consequently, each line represents a purple \textit{nominal} signal connection.}
    \label{fig:Hanselaar2022}
\end{figure}  

\subsection{High-level Safety Shell arbitration logic}
\label{section:highLevelSafetyShellLogic}
To introduce the Safety Shell, consider Fig. \ref{fig:Hanselaar2022}, which highlights three parallel AD channels capable of journey continuation, each with their own WM and MP functions, with \textit{Channel n} referring to the expandability of the architecture to more parallel channels.
Conceptually, the Safety Shell is based on the following ingredients: 
\begin{itemize}
    \item[a)] Given multiple heterogeneous channels available in the architecture, run-time determination of a relative preference of each channel through recent performance evaluation and initial design-time settings can be created.
    \item[b)] Risk computations can be executed through a safety test between all generated trajectories by the MPs and all available WMs.
\end{itemize}
Based on these two ingredients, the Safety Shell architecture is completed by an arbiter, which accumulates all risk and preference assessments, and selects a channel to control the vehicle at every evaluated timestep.
Inside the arbiter a fallback MP is available, which uses the accumulated WM data to generate an emergency trajectory to a safe state, in case none of the channels provides a sufficiently safe motion plan.
Sensors, not shown in Fig. \ref{fig:Hanselaar2022}, may be shared between some channels, but redundancy considerations can require at least some sensors {unique to different channels}~\cite{iso26262,iso21448}.

By exploiting multiple parallel journey continuing AD channels, the Safety Shell can overcome the limitations of simple fallback motion planners that cannot deal with complex scenarios \cite{Fruehling2019}.
Note that, as concluded in Section \ref{section:gap}, the evaluated architectures in Table \ref{table:overviewOfArchitectures} all sacrifice the availability of mission continuing functions for an increase in safety through the use of simple fallback MPs or motion envelop restrictions. 
As each of the parallel AD channels in the Safety Shell can possibly continue the journey, the availability is impacted less by switching to another channel.
For instance, even if one channel has a false positive observation (e.g., a ghost object is detected on the planned vehicle path), the journey-continuing nature of this channel allows it to go beyond simply slowing down and, instead, consider changing lanes or choosing an entirely different route.
Of course, one should try to avoid frequent switching between channels, as this impacts both the comfort of passengers and the predictability of the automated vehicle when viewed by other road users.
The arbiter algorithm used to select the channel has to balance all these aspects and is, therefore, instrumental in the success of the Safety Shell architecture.
We summarize these points through the following three rules:
\begin{enumerate}
    \item Select the most preferred channel, if safety and comfort allow it;
    \item Switch to a safe and sufficiently preferred channel, if the currently selected channel is predicted to create a dangerous situation;
    \item If the current channel is immediately dangerous and no safe alternative channel is available, then the emergency trajectory must be activated to avoid or at least mitigate unreasonable risk.
\end{enumerate}

\begin{figure}[h]
    \centering
    \includegraphics[width=7.2cm]{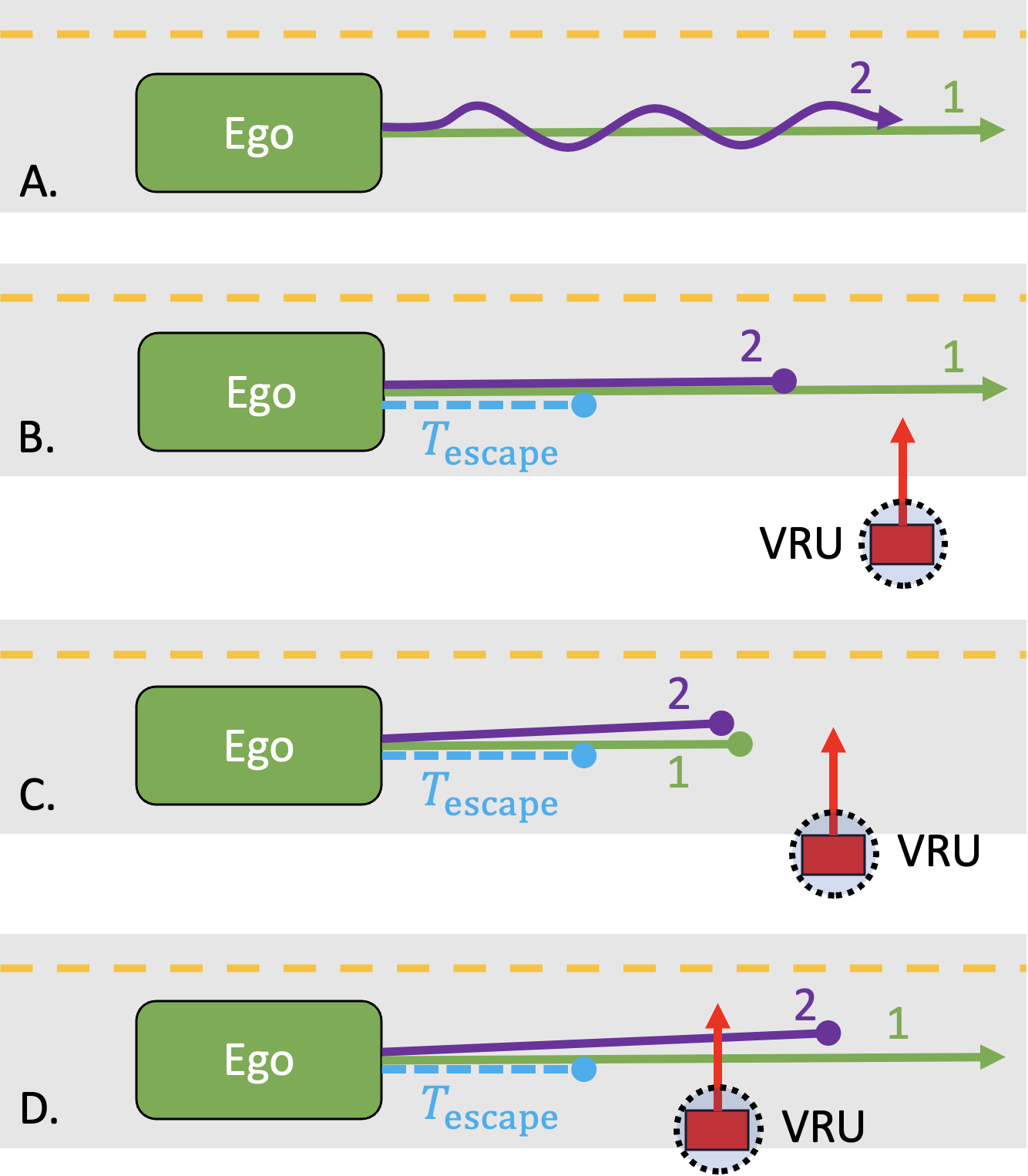}
    \caption{{Four example situations, with A. showing two safe channels of different quality, 
    B. showing a predicted dangerous situation along channel 1's trajectory,
    C. showing the possible result of giving channel 1 the benefit of the doubt and allowing it to create a safe trajectory too, and D. showing a possible situation where both channels do not provide a safe trajectory, and the last safe intervention time approaches, visualized by the fallback trajectory} $T_{\mathrm{escape}}$.}
    \label{fig:progressiveExamples}
\end{figure}

{To illustrate these rules, we propose four examples scenarios shown in Fig.} \ref{fig:progressiveExamples}. 
{In Fig.} \ref{fig:progressiveExamples}.A {the Ego vehicle, fitted with the Safety Shell, travels along a straight road without obstacles.
If both channels provide a safe trajectory, but channel 1 proposes a more preferred (e.g., more comfortable) trajectory, then a switch to channel 1 is desired.}
{This behaviour is encoded within rule 1.}
{Now consider a similar straight-road scenario, in which a pedestrian is predicted to cross the road. 
Let us consider the situation in which channel 1 plans a dangerous trajectory with respect to the pedestrian for some reason (e.g., no object detection, different predicted motion, insufficiency in the MP), while channel 2 plans to slow down smoothly. 
This situation with the planned trajectories, as well as a planned basic fallback manoeuvre denoted as} $T_{\mathrm{escape}}${, are drawn in Fig.} \ref{fig:progressiveExamples}.B.
{Given that channel 1 has been established as more preferred, it might be sensible to maintain our selection of channel 1 for a short amount of time, to allow channel 1 to respond appropriately, for example as drawn in Fig.} \ref{fig:progressiveExamples}.C.
{However, if channel 1 does not correct itself and channel 2 still provides a safe alternative, a switch to channel 2 is preferred. 
Certainly this is more preferred than an immediate switch to the} $T_{\mathrm{escape}}$ {trajectory, given that this will be much less comfortable and will interrupt the availability of autonomous functionality.
This balance between preference and safety is made in rule 2.}
{If, for some reason, both channels 1 and 2 do not create safe trajectories with respect to detected dangerous obstacles, the vehicle must attempt to reduce the risks.
In this situation, shown in Fig.} \ref{fig:progressiveExamples}.D{, maintaining the safety of the vehicle is the only relevant consideration. 
If there is still some safety margin between the current moment and the last safe time for the escape trajectory activation, then we may opt to wait to see if one of the channels may still be able to create a safe trajectory. 
However, safety is key in this case, and the fallback response is therefore activated at or close to this \textit{last safe intervention time}.
This consideration is covered by Rule 3.}

\subsection{High-level computation of arbitration logic}
\label{section:HighLevelMathSafetyShell}
{As unknown FIs may occur at any time during operation,
the Safety Shell must be run at a regular and sufficiently small periodic time interval $\Delta_{\mathrm{s}}$ to match sensor update periods (typically in the order of 0.2, 0.1 or 0.05 [s]).
This allows the Safety Shell to use the latest available information for FI detection and mitigation.}
The evaluation of the safety shell at time $t= k\Delta_{\mathrm{s}}$ is indicated by discrete time index $k \in \mathbb{N}$.
At each discrete time $k$ we approximate the safety of a planned trajectory in comparison to the observed and predicted surroundings based on all WMs, as well as differentiate acceptable risk from unacceptable risk, as we explain further below (Sections \ref{section:RiskCalculation} to \ref{section:ArbitrationStrategy}).

The three rules {given above} translate to the following high-level conceptual setup of the Safety Shell, where $C(k)~\in~\{1,2,...,n_{\mathrm{c}},T_{\mathrm{escape}}{(k)}\}$ represents the channel choice at discrete time $k$, for a system with $n_{\mathrm{c}}$ number of channels and integrated fallback trajectory $T_{\mathrm{escape}}{(k)}$, {calculated for the situation at time $k$}. 
Relating to ingredients a) and b), we assume that for each channel $i \in \overline{n}_{\mathrm{c}}:=\{1,2,...,n_{\mathrm{c}} \}$ a preference indicator $\pi_i(k) \in \mathbb{R}_{\geq0}$ and a risk-dependent safety indicator $\zeta_i(k) \in \mathbb{R}_{\geq0}$ are available at time $k\in \mathbb{N}$.
To emphasize, these quantities are time-varying and {thus} depend on $k$.
These conditions and definitions will be substantiated in later subsections. 
We assume that the channel selected at the prior timestep $k-1$ is given by index $j \in \overline{n}_{\mathrm{c}}$, i.e.,
\begin{equation}
    C(k-1) = j
    \label{eq:preAssignedChannelChoice}
\end{equation}
with the channel selection at initialization $C(-1)$ set to the most preferred channel available.
Next, we determine at which point in the past (if any) a channel switch occurred, i.e., 
\begin{equation}
    s(k):= \max\{l \in \mathbb{N}_{<k}| C(l) \neq C(l-1)\}
    \label{eq:lastChannelSwitch}
\end{equation}
where we use the convention that $\max(\varnothing)=0$. 
Next, we assess if sufficient time has passed since the last channel switch to allow a purely preference-based channel switch, through
\begin{equation}
    k-s(k) \geq q
    \label{eq:lastChannelSwitchSufficientlyLongAgo}
\end{equation}
where $q\in \mathbb{N}_{\geq 1}$ is a to-be-set parameter in the Safety Shell. 
The check \eqref{eq:lastChannelSwitchSufficientlyLongAgo} is used to avoid too frequent channel switches that are purely based on preference (which may be uncomfortable), but will not impede safety-based channel switches.
If \eqref{eq:lastChannelSwitchSufficientlyLongAgo} is true, we assess if there are more preferred and safe channels available via
\begin{equation}
    \{  i \in \overline{n}_{\mathrm{c}} \ | \
    {\zeta}_{i}(k) \geq \underline{\zeta}_{\mathrm{safe}} \land \pi_{i}(k) > \pi_{j}(k)
    \} \neq \varnothing
    \label{eq:SetOfMorePreferredSafeChannels}
\end{equation}
where ${\zeta}_{i}(k) \geq \underline{\zeta}_{\mathrm{safe}}$ is a sufficient condition of (a large degree of) safety of channel $i \in \overline{n}_{\mathrm{c}}$ based on the risk-dependent safety indicator ${\zeta}_{i}(k)$ and a minimum safety level $\underline{\zeta}_{\mathrm{safe}}$ and
$\pi_i(k)$ represents, as mentioned above, the (time-varying) preference score of each channel $i \in \overline{n}_{\mathrm{c}}$, at the discrete time $k\in\mathbb{N}$.
Consequently, $\pi_i(k) > \pi_j(k)$ restricts the set of suitable channels in \eqref{eq:SetOfMorePreferredSafeChannels} to only those that are preferred more than the currently selected channel $j$.
If both \eqref{eq:lastChannelSwitchSufficientlyLongAgo} and \eqref{eq:SetOfMorePreferredSafeChannels} are true, we select the most preferred safe channel via
\begin{equation}
    C(k):= \underset{i\in\overline{n}_{\mathrm{c}}}{\mathrm{argmax}} \{ \pi_i(k)|{\zeta}_{i}(k) \geq \underline{\zeta}_{\mathrm{safe}} \land \pi_i(k) > \pi_j(k)\}
    \label{eq:chooseMostPreferredChannel}
\end{equation}
Together, \eqref{eq:lastChannelSwitch} through \eqref{eq:chooseMostPreferredChannel} form the implementation of rule 1. 
If \eqref{eq:lastChannelSwitchSufficientlyLongAgo} or \eqref{eq:SetOfMorePreferredSafeChannels} (or both) are false, 
we assess if a channel switch is required to reduce exposure to {significant} risk predicted along the trajectory of the currently active channel $j$.
A significant risk predicted to occur within a limited time due to the planned trajectory $T_j$ of channel $j$ results in ${\zeta}_{j}(k) < \underline{\zeta}_{\mathrm{safe}}$.
However, ${\zeta}_{j}(k) < \underline{\zeta}_{\mathrm{safe}}$ does not mean an \textit{immediate} intervention is required.
For instance, the identified risk may be predicted to occur some time in the future, providing the automated vehicle with plenty of time to respond appropriately. 
In these scenarios, we re-use the safety indicator ${\zeta}_j(k)$ to restrict the set of channels considered as suitable alternatives to those channels with sufficiently high preference scores.
This selection procedure is indicated via
\begin{equation}
    \{ i \in \overline{n}_{\mathrm{c}} \ | \ 
    {\zeta}_{i}(k) \geq \underline{\zeta}_{\mathrm{safe}} \land \pi_i(k) \geq {\zeta}_j(k)
    \} \neq \varnothing
    \label{eq:SetOfSafetySwitchChannels}
\end{equation}
The comparison between preference $\pi_i(k)$ and risk indicator $\zeta_i(k)$ in \eqref{eq:SetOfSafetySwitchChannels} may appear strange at first, but will be elaborated on more in Section \ref{section:PreferenceOrderAndSwitchTime} below. 
By assigning higher preference scores $\pi_i(k)$ to channels that aim to continue the journey instead of reverting to a safe-stop, 
{the Safety Shell} will first consider {selecting these} availability-maintaining alternative channels {as a way} to reduce risk. 
If risks are predicted to occur far enough in the future for the currently selected channel $j$, i.e., $\zeta_j(k)$ is relatively large, such that none of the channels has a sufficiently high preference score (i.e., $\pi_i(k)<{\zeta}_j(k)$, for all $i \in \overline{n}_{\mathrm{c}}$), then the limits of the set of channels in \eqref{eq:SetOfSafetySwitchChannels} allows the vehicle to continue with the prior selection, i.e., $C(k):= j$.
{This allows} e.g., false-positive risks to be corrected in the next Safety Shell evaluation at $k+1$ without requiring a channel switch.
If \eqref{eq:SetOfSafetySwitchChannels} is true, {then} channel $j$ exposes the vehicle to risk and at least one suitable alternative channel is available. 
{Of those, the} most preferred is identified via
\begin{equation}
    C(k):= \underset{i\in \overline{n}_{\mathrm{c}}}{\mathrm{argmax}} \{ \pi_i(k)|{\zeta}_{i}(k) \geq \underline{\zeta}_{\mathrm{safe}} \land \pi_i(k) \geq {\zeta}_j(k)\}
    \label{eq:safetyDrivenPreferredChoice}
\end{equation}
Equations \eqref{eq:SetOfSafetySwitchChannels} and \eqref{eq:safetyDrivenPreferredChoice} represent the implementation of rule 2.

If both \eqref{eq:SetOfMorePreferredSafeChannels} and \eqref{eq:SetOfSafetySwitchChannels} are false, we assess if we can either continue with the currently selected channel or if we must use an emergency trajectory, through
\begin{equation}
 C(k):=
    \begin{cases}
        T_{\mathrm{escape}}{(k)} & \text{if } {\zeta}_j(k) \leq \overline{\zeta}_{\mathrm{unsafe}} \\
        j & \text{if } {\zeta}_j(k) > \overline{\zeta}_{\mathrm{unsafe}} \\
    \end{cases}
    \label{eq:escapeConsideration}
\end{equation}
where $\overline{\zeta}_{\mathrm{unsafe}} < \underline{\zeta}_{\mathrm{safe}} $ represents an immediate danger threshold
and $T_{\mathrm{escape}}{(k)}$ represents the fallback motion planner's escape trajectory {at discrete time $k \in \mathbb{N}$}.
Finally, \eqref{eq:escapeConsideration} forms the implementation of rule 3.

Together, equations \eqref{eq:preAssignedChannelChoice} through \eqref{eq:escapeConsideration} represent the translation of the two ingredients and three rules into the high-level formulation of the Safety Shell arbitrator using safety indicators $\zeta_i(k)$ and preference indicators $\pi_i(k)$ for each channel $i\in \overline{n}_{\mathrm{c}}$.
Section \ref{section:RiskCalculation} through  \ref{section:SufficientlySafe} will now detail the implementations regarding risk, a time-based evaluation of danger and the interpretation of sufficient safety.
In these subsections, we will specify 
$\pi_i(k)$, ${\zeta}_{j}(k)$, $\underline{\zeta}_{\mathrm{safe}}$, $\overline{\zeta}_{\mathrm{unsafe}}$ and $T_{\mathrm{escape}}{(k)}$.
Identifying which channels can be considered through their preference scores is explained in Section \ref{section:PreferenceOrderAndSwitchTime}.
The arbitration strategy introduced above is updated with the implementation  details of Sections \ref{section:RiskCalculation} to \ref{section:PreferenceOrderAndSwitchTime} in Section \ref{section:ArbitrationStrategy}.

\subsection{Risk calculations}
\label{section:RiskCalculation}
To quantify the {safety} of a trajectory, this work uses a relative risk calculation.
We use the general definition of risk from \cite{iso26262} as the combination of the predicted probability of adverse event $\mathrm{E}\in \{ 1,2,...,n_{\mathrm{events}}\}$ occurring combined with the expected severity of that event.
An adverse event $\mathrm{E}$ is not restricted to collisions and may also refer to, e.g., loss-of-control, rule-restriction-violation or out-of-lane events.
At each risk function call at discrete time $k$ by the Safety Shell arbiter, risk is evaluated over a prediction time horizon $\tau_{\mathrm{H}}\Delta_p$ into the future. In particular, we take $\tau \in \mathbb{N}:=\{0,1,2,...,\tau_{\mathrm{H}} \}$ as the predicted discrete time beyond the current time $t=k\Delta_s$, with the associated constant prediction time step size of $\Delta_{\mathrm{p}}$~[s].
A quantity, e.g., risk, evaluated at the discrete time $k\in \mathbb{N}$, 
concerning the future predicted discrete moment $\tau \in \mathbb{N}$, refers then to the time interval
$\left[k\Delta_{\mathrm{s}}+ \tau \Delta_{\mathrm{p}},\ k\Delta_{\mathrm{s}}+(\tau+1) \Delta_{\mathrm{p}} \right)$.
The prediction time step $\Delta_{\mathrm{p}}$ may differ in size from the evaluation time step $\Delta_{\mathrm{s}}$. 
{
To calculate the risks that inform channel choice, we compare trajectories to WMs. 
As trajectories and WMs are updated regularly by channels, they are time-varying.
A trajectory from channel $j$ will contain the planned consecutive states of the vehicle, via
\begin{equation}
    T_j(k):=\{ T_j(k,0),T_j(k,1),T_j(k,2),...,T_j(k,\tau_{\mathrm{H}}) \}
\end{equation}
where the notation $T_j(k,\tau)$ is used to represent the planned vehicle state at $k\Delta_{\mathrm{s}}+ \tau \Delta_{\mathrm{p}}$.
WM$_i(k)$ will contain similar predicted trajectories for all relevant object identified at time $k$ and may contain various observations that do not change over the prediction horizon, e.g., road boundaries, weather conditions.
The implementation of this notation is shown in the example in Section \ref{section:Results} and specifically Fig. \ref{fig:exampleWMXIncorrectLocationSaS2} and \ref{fig:exampleRiskFigureIncorrectLocationSaS2}.
To lighten the notation we have dropped the $k,\tau$ dependencies in $T_j$ and WM$_i$ where they do not offer extra insights.
Similarly, we do not explicitly denote the dependency on $k$ when we use the variable $\tau$, as we assume $k$ to be clear from the context.}
The risk $ R_{ij}(\tau)$ is {subsequently} calculated as
\begin{equation}
    R_{ij}(\tau) = \sum_{\mathrm{E}=1}^{n_{\mathrm{events}}} P_{\mathrm{E},ij}(\tau) S_{\mathrm{E},ij}(\tau)
    \label{eq:risk_simplified}
\end{equation}
{where $i \in \overline{n}_{\mathrm{c}}$ is the channel index of the world model $\mathrm{WM}_i$ that we use to assess the risk of trajectory $T_j$ from channel $j \in \overline{n}_{\mathrm{c}}$.}
The term $P_{\mathrm{E},ij}(\tau)$ represents the probability of adverse event $\mathrm{E} \in \{1,2,...,n_{\mathrm{events}} \}$ to occur in the time interval $\left[k\Delta_{\mathrm{s}}+ \tau \Delta_{\mathrm{p}},\ k\Delta_{\mathrm{s}}+(\tau+1) \Delta_{\mathrm{p}} \right)$, given the trajectory $T_j$ and world model $\mathrm{WM}_i$ {at} discrete time $k$.
Following e.g., \cite{Damerow2015risk,Eggert2014risk}, we use a map from conventional safety indicator values (e.g., Time To Collision \cite{tamke2011flexible}, Post Encroachment Time \cite{Horst1990PET} and minimum safe distance) to an approximate  probability of an adverse event.
For further details on setting up the probability calculations, we refer to Appendix \ref{appendix:risk}.

The term $S_{\mathrm{E},ij}(\tau)$ in \eqref{eq:risk_simplified} refers to the severity of the same adverse event E in the time interval $\left[k\Delta_{\mathrm{s}}+ \tau \Delta_{\mathrm{p}},\ k\Delta_{\mathrm{s}}+(\tau+1) \Delta_{\mathrm{p}} \right)$, again based on WM$_i$ and $T_j$.
To create suitable severity estimations we used \cite{Richards2010} to relate predicted impact speed to injury severity outcome probability.
The curves found by \cite{Richards2010} range from zero probability of severe injury or death at near-zero speeds of collision to unity at higher speeds of collision.
To ensure that all collisions are penalised to some extent in the Safety Shell, we raised the severity of collisions with vehicles and Vulnerable Road Users (VRUs) at all speeds by 1. 
Consequently, the severity now ranges from 1 to 2 for VRU and vehicle objects, though the rate at which they increase as a function of the predicted impact speed varies per type of object.
The calculation of severity is shown in Appendix \ref{appendix:severityScore}.
Other suitable risk calculation methods can be used as well in the Safety Shell, indicating its flexibility. 

An experimentally determined risk thresholds value $R_{\mathrm{Threshold}}$ is set as per \cite{Damerow2015risk,Eggert2014risk}, which specifies unreasonable risks via
\begin{equation}
    R_{ij}(\tau)\ge R_{\mathrm{Threshold}}
    \label{eq:RiskThresholdExceedence}
\end{equation}
Using \eqref{eq:RiskThresholdExceedence}, we can identify the first moment a trajectory exceeds the unreasonable risk threshold via
\begin{equation}
\begin{aligned}
    \tau_{\mathrm{U},j}{(k)}:={\mathrm{min}}\{ \tau \in \{  0,1,2,...,\tau_{\mathrm{H}}\}\ |\ \\
    R_{ij}(\tau)\ge R_{\mathrm{Threshold}} \ \ \\
     \mathrm{for}\ \mathrm{some} \ i \in \overline{n}_{\mathrm{c}} \}
\end{aligned}
\label{eq:firstUnreasonableRisk}
\end{equation}

If $R_{ij}(\tau)< R_{\mathrm{Threshold}}$ for a tested trajectory $T_j$, for all $\tau \in \{0,1,2,...,\tau_{\mathrm{H}}\}$ according to all WM$_i$ with $i \in \overline{n}_{\mathrm{c}}$ (i.e., the sets in the right-hand side of \eqref{eq:firstUnreasonableRisk} are all empty) we set $\tau_{\mathrm{U},j}{(k)}:=\infty$.
Large values for $\tau_{\mathrm{U},j}{(k)}$ indicate a large degree of safety for trajectory $T_j$.

\subsection{Last safe intervention time}
\label{section:LastSafeInterventionTime}
The last safe intervention time $\tau_{\mathrm{L},j}(k)$, $k\in \mathbb{N}$, is the amount of time (if any) left until the automated vehicle must intervene with an emergency response to avoid all predicted unreasonable risk, if it continues to follow the trajectory $T_j$ up to future time $\tau_{\mathrm{L},j}(k)$, similar to the \textit{plan b} approach in~\cite{Damerow2015risk} or \textit{time to response} from \cite{tamke2011flexible}.
{In Section} \ref{section:highLevelSafetyShellLogic} {we already alluded to this concept in the shown example accompanying Fig.} \ref{fig:progressiveExamples}.
Formally, we define the last safe intervention time via
\begin{equation}
\label{eq:LSI_calc_generalized}
\begin{aligned}
    \tau_{\mathrm{L},j}(k):=\mathrm{max} \{\theta\in\{0,1,...,\tau_{\mathrm{U},j}{(k)}-1\}  \ | \ \ \  \\
      R_{ij}^{\theta}(\tau) < R_{\mathrm{Threshold}}, \ \ \\
      \forall i\in\overline{n}_{\mathrm{c}}, \ \ \\
      \forall \tau \in \{0,1,...,\tau_{\mathrm{H}}\} \}
\end{aligned}
\end{equation}
where $R_{ij}^{\theta}(\tau)$ is the risk re-calculated at time $k$ via \eqref{eq:risk_simplified} over the full prediction time horizon $\tau \in \{0,1,...,\tau_{\mathrm{H}}\}$, based on $\mathrm{WM}_i$, $i \in \overline{n}_{\mathrm{c}}$, with the tested trajectory $T_j^{\theta}$ defined by
\begin{equation}
  T_j^{\theta}(\tau) :=
    \begin{cases}
      T_j(\tau) & \text{for all $\tau < \theta$}\\
      T_{\mathrm{escape},j}(\tau) & \text{for all $\tau \geq \theta$}\\
    \end{cases}   
\label{eq:T_combined_defintion}
\end{equation}
for $\theta\in\{0,1,...,\tau_{\mathrm{U},j}(k)-1\}$. 
{In other words, the states $T_j^{\theta}(\tau)$ in trajectory $T_j^{\theta}$ are equal to the states $T_j(\tau)$ in trajectory $T_j$ up to but not including $\tau = {\theta}$, while the following states $T_j^{\theta}(\tau)$ are set according to $T_{\mathrm{escape},j}(\tau)$.}
If no suitable intervention exists that eliminates all unreasonable risk, i.e., if no $\theta\in\{0,1,...,\tau_{\mathrm{U},j}{(k)}-1\}$ creates a combined trajectory $T_j^{\theta}$ through \eqref{eq:T_combined_defintion} that satisfies $R_{ij}^{\theta}(\tau) < R_{\mathrm{Threshold}} $ in \eqref{eq:LSI_calc_generalized}, or if $\tau_{\mathrm{U},j}{(k)}=0$ (unreasonable risk is immediate), we set $\tau_{\mathrm{L},j}(k):=0$ for trajectory $T_j$.
If $\tau_{\mathrm{U},j}{(k)}=\infty$ then we assign $\tau_{\mathrm{L},j}(k):=\infty$.
The escape trajectory $T_{\mathrm{escape},j}$ used in \eqref{eq:T_combined_defintion} is defined as a trajectory that uses the maximum allowable capacity of the vehicle to reduce the predicted exposure to risk.
As a first proof-of-concept implementation of $T_{\mathrm{escape},j}$, an AEB manoeuvre is used, similar to the safety procedures from, e.g., \cite{Nister2019,Shalev-Shwartz2017}, though other manoeuvres can also be used.
Due to the simple nature of this escape trajectory, the lateral control of $T_{\mathrm{escape},j}$ is based on the intended path of $T_j$, hence its dependency on channel index $j$.
This definition of the last safe intervention time $\tau_{\mathrm{L},j}(k)$ forms a proportional measure to the urgency of an intervention.
We use this as the risk-based safety indicator $\zeta_i(k)$ as used in Section \ref{section:HighLevelMathSafetyShell}, i.e.,
\begin{equation}
     {\zeta}_j(k):= \tau_{\mathrm{L},j}(k) 
     \label{eq:EquivalentRiskIndexLSI}
\end{equation}

\subsection{Quantifying safety}
\label{section:SufficientlySafe}
\begin{figure}
    \centering
    \includegraphics[width=6.5cm]{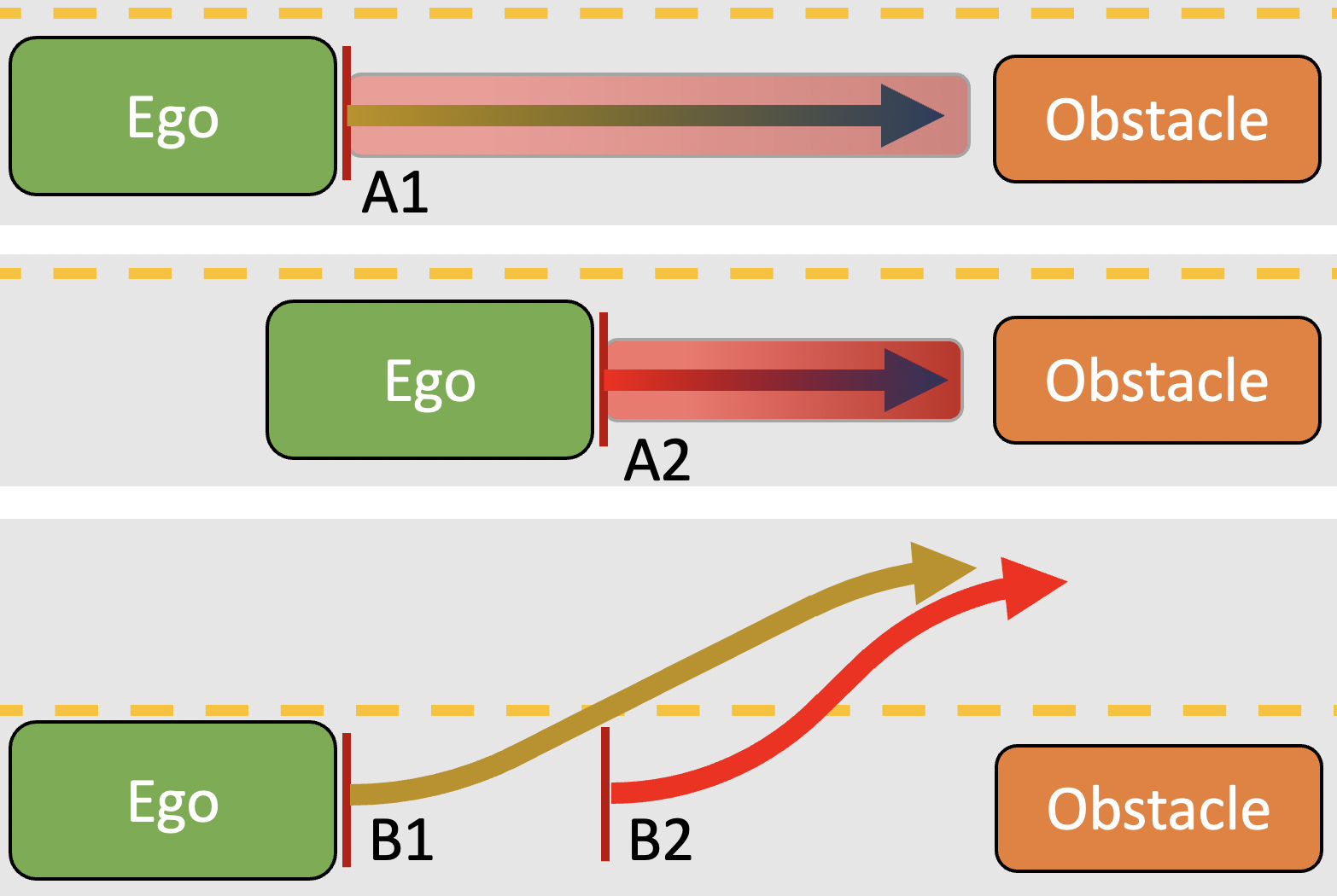}
    \caption{Illustration of the difference in required intervention to prevent unreasonable risk due to a stationary object, for 1D motion space (A1 and A2) and 2D motion space (B1 and B2 respectively).}
    \label{fig:InterventionComparison}
    \vspace{-7pt}
\end{figure}

The arbitration logic introduced in Section \ref{section:HighLevelMathSafetyShell} relies on quantified safety levels.
The prior subsection introduced the substitution of $\tau_{\mathrm{L},j}(k)$ for ${\zeta}_j(k)$ in \eqref{eq:EquivalentRiskIndexLSI}.
Here, we will introduce a substitution for thresholds defining safe $ \underline{\zeta}_{\mathrm{safe}} $ and unsafe $\overline{\zeta}_{\mathrm{unsafe}}$ trajectories, based again on $\tau_{\mathrm{L},j}(k)$.
We illustrate the process of quantifying safety using the following simplified considerations.
Generally a risk-reducing trajectory started prior to $\tau_{\mathrm{L},j}(k)$ will require less than the maximum manoeuvring capacity to avoid all predicted unreasonable risk.
As illustrated in the 1D example in the top of Fig.~\ref{fig:InterventionComparison}, switching from the planned dangerous trajectory to a decelerating alternative trajectory results in an intuitively smoother intervention, if the switch is performed sooner (see A1 vs A2 in Fig.~\ref{fig:InterventionComparison}).
Similarly, a 2D trajectory can be smoothed considerably, if more time is available to evade the obstacle that causes risk (see B1 vs B2 in Fig.~\ref{fig:InterventionComparison}).
Reversing the argument, we pose that a trajectory is sufficiently safe if: 1)~there is no required intervention ($\tau_{\mathrm{L},j}=\infty$), or 2)~if the required intervention, when immediately implemented, only requires relatively comfortable accelerations to avoid all unreasonable risk.
This limit can be expressed either as an upper limit of the vehicle capacity required for the escape trajectory (e.g., 25\% of available brake deceleration capacity, in the case of an AEB-based escape trajectory), or as an analogous minimum time until the last safe intervention time, e.g., $\Delta_{\mathrm{p}} \tau_{\mathrm{L},j}(k)  \geq t_{\mathrm{suff}}$ [s].
The capacity threshold or value of $t_{\mathrm{suff}}$ may be determined through expert assessment or experimentation.
To facilitate discrete-time implementation, $t_{\mathrm{suff}}$ is expressed as
$t_{\mathrm{suff}} = \tau_{\mathrm{suff}} \Delta_{\mathrm{p}}$.
Using this, we set $\underline{\zeta}_{\mathrm{safe}}$ in \eqref{eq:safetyDrivenPreferredChoice} as
\begin{equation}
     \underline{\zeta}_{\mathrm{safe}}:=\tau_{\mathrm{suff}}
    \label{eq:sufficientlySafe}
\end{equation}
We suggest $\tau_{\mathrm{suff}}\Delta_p \geq1.5$ [s], depending on the operational design domain.

Next, we substitute the immediate danger threshold ${\zeta}_{j}\leq \overline{\zeta}_{\mathrm{unsafe}}$ used in \eqref{eq:escapeConsideration} with
\begin{equation}
    \overline{\zeta}_{\mathrm{unsafe}}:=\tau_{\mathrm{immediate}}
    \label{eq:ImmediateDangerThreshold}
\end{equation}
with $t_{\mathrm{immediate}}=\tau_{\mathrm{Immediate}}\Delta_p$ in the order of 0.1 to 0.3 [s].
The safety margin $t_{\mathrm{immediate}}$ is used to ensure that actuation delays as a result of vehicle characteristics are compensated through timely activation of the escape trajectory $T_{\mathrm{escape},j}$.
To avoid cyclic logic, $\tau_{\mathrm{suff}}>\tau_{\mathrm{Immediate}}$ must be true.

As the time thresholds generalize easily over any type of escape trajectory and WM and MP functions have limited prediction horizons, we have opted to define all the thresholds in time.
This allows predictable behaviour in various scenarios requiring arbitration decisions.

\subsection{Preference order and consideration times}
\label{section:PreferenceOrderAndSwitchTime}
If $\tau_{\mathrm{immediage}}<\tau_{\mathrm{L},j}(k)<\tau_{\mathrm{safe}}$, then the selected channel $j$ is no longer sufficiently safe, yet it is not immediately dangerous.
In this case the arbitration strategy considers alternative channels to the currently selected channel, as introduced in \eqref{eq:SetOfSafetySwitchChannels} through \eqref{eq:safetyDrivenPreferredChoice}.
As indicated in the prior subsection, intervening or switching sooner decreases the intervention severity.
However, as mentioned at the start of Section~\ref{section:SafetyShellDesignPattern}, frequent channel switches are likely very uncomfortable. 
Consequently, switching as soon as possible will likely not result in a smooth experience in the automated vehicle, which is why \eqref{eq:lastChannelSwitchSufficientlyLongAgo} was introduced and we now introduce preference-order based channel selection in face of future risks.
We will now define $\pi_i(k)$ used in \eqref{eq:SetOfMorePreferredSafeChannels}-\eqref{eq:safetyDrivenPreferredChoice}, via a so-called consideration time $t_{\mathrm{C},i}(k) = \tau_{\mathrm{C},i}(k)\Delta_{\mathrm{p}}$, $i\in \overline{n}_{\mathrm{c}}$, explained below.
Following \eqref{eq:SetOfSafetySwitchChannels}, channel $i$ is only considered if $\pi_i(k) \geq {\zeta}_j(k)$. 
This preliminary condition is now updated to
\begin{equation}
    \pi_i(k) \geq {\zeta}_j(k) \Leftrightarrow \tau_{\mathrm{C},i}(k)\geq \tau_{\mathrm{L},j}(k)
    \label{eq:considerationTime}
\end{equation}
The consideration time $\tau_{\mathrm{C},i}(k)$ is a time-varying preference score and a function of the consideration time set at design time, $\tau_{\mathrm{C},i}^*$.
Unfortunately, there is no formal rule to tune {these} initial parameters. 
E.g. if you find that in a two-channel system one of the channels is prone to cause false-positive risk estimations and provides less comfort, setting its consideration time low will ensure that it is not immediately selected and this channel has time to correct its false-positive risk estimations.
On the other hand if both channels are considered good alternatives that rarely cause false-positive risk estimations, you may choose to set the consideration {closer to one another}. 
{This idea, to balance preference and safety, was introduced through rule 2 in Section }\ref{section:highLevelSafetyShellLogic}, {see also the accompanying example in Fig. }\ref{fig:progressiveExamples}{.}
To create an initial estimate of consideration times we use brake deceleration targets of the included channels as a substitute of comfort and apply them to the simple 1D case shown at the top of Fig. \ref{fig:InterventionComparison}.
We determine how much time prior to the $\tau_{\mathrm{L},j}$ we need to switch to another channel $i\neq j, i \in \overline{{n}}_{\mathrm{c}}$, to attain a desired level of comfort.
E.g., if stationary objects are reasonable to expect on path up to ego velocities of $v=20$ m/s, we can use the difference between the maximum allowable brake deceleration $a_{\mathrm{L}}$ (used for the calculation of the Last Safe Intervention Time) and the desired magnitude of brake intervention required by a channel $a_i$, to estimate the when we need to switch to the channel to not necessitate more severe braking than $a_i$. 
To do so, we compute the difference of the expected braking distances of the escape intervention and the assumed channel intervention, and divide by the time taken to cover that distance, via
\begin{equation}
    \tau_{\mathrm{C},i}^* = \frac{1}{v}\left(\frac{v^2}{2a_i} - \frac{v^2}{2a_L} \right)
\end{equation}
Setting  $a_{\mathrm{L}} = 8$ m/s$^2$ and $a_1=3.5$ m/s$^2$ returns $\tau_{\mathrm{C},i}^* \approx 1.6$ s, while setting $a_2=4.5$ m/s$^2$ returns $\tau_{\mathrm{C},i}^* \approx 1.0$ s. 

As concluded in Section \ref{section:stateoftheart}, no design-time FI elimination method is complete. 
Consequently, each ADS will contain some unknown FIs at release.
Therefore, it is possible that the AD channel assumed to be best during the automated vehicle design-time performs poorly in specific scenarios.
To allow the arbitration logic to select the best performing channel, we track the recent channel performance, via
\begin{equation}
    \tau_{\mathrm{C},i}(k) = \tau^{*}_{\mathrm{C},i} \left( \frac{1}{1+\rho g_i(k)} \right)
    \label{eq:ConsiderationTimeCalculation}
\end{equation}
with $g_i(k)$ the number of occurrences of $\tau_{\mathrm{L},i} < \tau_{\mathrm{suff}}$ over a rolling time window $\{k-k_r,...,k\}$ of length $k_r \in \mathbb{N}^{{+}}$ of operation and $\rho$ an experimentally determined scaling factor.
By tracking early indications of reduced safety, $g_i(k)$ can serve as a Safety Performance Indicator~\cite{Koopman2020c}. 
Other performance tracking metrics can also be used instead of or in addition to \eqref{eq:ConsiderationTimeCalculation} to adjust the consideration times.

Using this definition of ${\pi}_i(k):=\tau_{\mathrm{C},i}(k)$ and the suggested tuning procedure, the comparison between preference and last safe intervention time is both possible and appropriate. 

\subsection{Final arbitration strategy}
\label{section:ArbitrationStrategy}
With all elements regarding last safe intervention time, consideration time and sufficient and immediate thresholds explained, we can rewrite \eqref{eq:SetOfMorePreferredSafeChannels} through \eqref{eq:escapeConsideration} to include them.
Starting with \eqref{eq:SetOfMorePreferredSafeChannels}, the substitutions of ${\zeta}_{j}(k) \geq \underline{\zeta}_{\mathrm{safe}} \Leftrightarrow \tau_{\mathrm{L},j}(k)\geq \tau_{\mathrm{suff}}$ \eqref{eq:sufficientlySafe} and $\tau_{\mathrm{C},i}(k)$ for $\pi_i(k)$ result in
\begin{equation}
    \{ i \in \overline{n}_{\mathrm{c}} \ | \ 
    \tau_{\mathrm{L},i}(k)\geq\tau_{\mathrm{suff}}\land \ \tau_{\mathrm{C},i}(k) > \tau_{\mathrm{C},j}(k) \} \neq \varnothing
    \label{eq:SetOfMorePreferredSafeChannels2}
\end{equation}
As introduced in Section \ref{section:HighLevelMathSafetyShell}, if \eqref{eq:lastChannelSwitchSufficientlyLongAgo} and \eqref{eq:SetOfMorePreferredSafeChannels2} are both true we {implement rule 1 from Section} \ref{section:highLevelSafetyShellLogic} and select the more preferred channel to continue via
\begin{equation}
    \label{eq:chooseMostPreferredChannel2}
    \begin{aligned}
        C(k):= \quad& \underset{i\in\overline{n}_{\mathrm{c}}}{\mathrm{argmax}} \{\tau_{\mathrm{C},i}(k) \ |\\
        \quad& \ \ \ \ \tau_{\mathrm{L},i}(k)>\tau_{\mathrm{suff}} \ \ \land \tau_{\mathrm{C},i}(k) > \tau_{\mathrm{C},j}(k)\}
    \end{aligned}
\end{equation}

If either \eqref{eq:lastChannelSwitchSufficientlyLongAgo} or \eqref{eq:SetOfMorePreferredSafeChannels2} is false, we evaluate if {we should switch to a sufficiently preferred safe channel, implementing rule 2 from Section }\ref{section:highLevelSafetyShellLogic} {via} the rewritten version of \eqref{eq:SetOfSafetySwitchChannels}, namely
\begin{equation}
    \{  i \in \overline{n}_{\mathrm{c}} \ | \ 
    \tau_{\mathrm{L},i}(k)\geq\tau_{\mathrm{suff}}\land \ \tau_{\mathrm{C},i}(k) \geq \tau_{\mathrm{L},j}(k) \} \neq \varnothing
    \label{eq:SetOfSafetySwitchChannels2}
\end{equation}
where we also substitute $\pi_i(k) \geq {\zeta}_j(k) \Leftrightarrow \tau_{\mathrm{C},i}(k)\geq \tau_{\mathrm{L},j}(k)$ \eqref{eq:considerationTime}.
If \eqref{eq:SetOfSafetySwitchChannels2} is true, we evaluate the best channel for the safety-driven intervention (previously \eqref{eq:safetyDrivenPreferredChoice}), via
\begin{equation}
    \label{eq:safetyDrivenPreferredChoice2}
    \begin{aligned}
        C(k):= \quad&\underset{i\in\overline{n}_{\mathrm{c}}}{\mathrm{argmax}} \{ \tau_{\mathrm{C},i}(k) \ |\\
        \quad& \ \ \ \  \tau_{\mathrm{L},i}(k)>\tau_{\mathrm{suff}} \ \ \land 
        \tau_{\mathrm{C},i}(k) \geq \tau_{\mathrm{L},j}(k)\}
    \end{aligned}
\end{equation}

Because \eqref{eq:safetyDrivenPreferredChoice2} is not restrained by \eqref{eq:lastChannelSwitchSufficientlyLongAgo} to allow for safety-based switching, consideration times need to adhere to $\tau_{\mathrm{C},i}(k) < \tau_{\mathrm{suff}}$ for all $i = \overline{n}_{\mathrm{c}}$ and all $k\in \mathbb{N}$ to ensure that \eqref{eq:safetyDrivenPreferredChoice2} does not cause fast channel switching between two channels with close $\tau_{\mathrm{C},i}(k) > \tau_{\mathrm{suff}}$ values and safe trajectories.
Finally, we {implement rule 3 from Section} \ref{section:highLevelSafetyShellLogic} {by rewriting} \eqref{eq:escapeConsideration} to 
\begin{equation}
 C(k):=
    \begin{cases}
        T_{\mathrm{escape},h} & \text{if } \tau_{\mathrm{L},j}(k) \leq \tau_{\mathrm{immediate}} \\
        j & \text{if }\tau_{\mathrm{L},j}(k) > \tau_{\mathrm{immediate}}
    \end{cases}
    \label{eq:escapeConsideration2}
\end{equation}
where we substituted $ \tau_{\mathrm{L},j}(k)$ for ${\zeta}_j(k)$ and $\tau_{\mathrm{immediate}}$ for $\overline{\zeta}_{\mathrm{unsafe}}$, and $T_{\mathrm{escape},h}$ is the safest of the escape trajectories, based on the assumptions outlined in Section \ref{section:SufficientlySafe}. 
It is defined as
\begin{equation}
    h:= \underset{i\in\overline{n}_{\mathrm{c}}}{\mathrm{argmax}}\{ \tau_{\mathrm{L},i}(k)\}
    \label{eq:escapeConsideration3}
\end{equation}
Note, the escape trajectory selection process in \eqref{eq:escapeConsideration3} is a consequence of the way the current escape trajectory is enforced to follow the originally planned path. 
If more sophisticated escape trajectory planners are available, the dependency on different channels and therefore on subscript $h$ can be dropped in \eqref{eq:T_combined_defintion}, \eqref{eq:escapeConsideration2} and \eqref{eq:escapeConsideration3}.

The arbitration algorithm now uses the following steps: 

\vspace{5pt}
\fbox{
\begin{minipage}{7.8cm}
{\begin{enumerate}
    \item Extracting the prior time step channel selection \eqref{eq:preAssignedChannelChoice} and assert when a different channel was selected via \eqref{eq:lastChannelSwitch};
    \item If sufficient time passed for preference-based switches \eqref{eq:lastChannelSwitchSufficientlyLongAgo} and more preferred channels are available, i.e., \eqref{eq:SetOfMorePreferredSafeChannels2} holds, assess \eqref{eq:chooseMostPreferredChannel2} to select a \textit{more preferred and sufficiently safe channel};
    \item If the above did not result in a channel choice and at least one suitably preferred and safer alternative is available, i.e., \eqref{eq:SetOfSafetySwitchChannels2} holds, assess \eqref{eq:safetyDrivenPreferredChoice2} to perform a safety-driven switch to the \textit{most preferred of all sufficiently safe channels}; 
    \item If the above did not result in a channel choice, test if the current channel is not immediately dangerous, or if we have to switch to the safest escape trajectory via \eqref{eq:escapeConsideration2} and \eqref{eq:escapeConsideration3}.
\end{enumerate}}
\end{minipage}
}
\vspace{3pt}

Through this logic the arbiter ensures that not all unreasonable risk identified on the selected channel trajectory $T_j$ requires immediate channel switching or mitigation.
This provides channel $j$ an opportunity to resolve any temporary FIs that may cause this unreasonable risk in $T_j$.
Next, delaying a channel switch also allows FP risk assessments by other channels to be corrected in subsequent Safety Shell algorithm evaluations.
All of this reduces channel switching and, consequently, improves comfort. 
In combination with \eqref{eq:chooseMostPreferredChannel2}, this arbitration strategy leads to increasing usage of the most preferred channel.

{In summary, the key differences to other architectures listed in Table} \ref{table:overviewOfArchitectures} {are:}
\begin{enumerate}
    \item {The inclusion of sophisticated redundant channels allows for better evaluation of complex scenarios and increases the likelihood that the redundant channels are able to deal with the difficult and complex scenarios that stump the preferred channel.}
    \item {By first considering more preferred and mission-continuing channels in} \eqref{eq:safetyDrivenPreferredChoice2} {instead of choosing the channel that maximizes safety in complex circumstances, the number of sudden stops are reduced and the automated vehicle's comfort and its ability to reach the travel destination is increased, when compared to alternative architectures.}
    \item {By converging to the best AD channels through }\eqref{eq:ConsiderationTimeCalculation}, {the number of required fallback MP activations will reduce, compared to architectures that can only resort to safe stop mitigation strategies.}
\end{enumerate}
{The escape-trajectory and last-safe-intervention-time based reasoning ensures that safety is maintained, at least to the level of other evaluated architectures, even if no included channel is able to produce an acceptable trajectory.}
These assessments will be supported by the numerical simulations in the next section {and illustrated by a highlighted test}.

\section{Numerical simulations}
\label{section:NumericalSims}
To evaluate the relative performance of the Safety Shell versus a selection of the architectures introduced in Section \ref{section:stateoftheart} we perform numerical simulations. 
These are created using the Automated Driving Toolbox from Matlab~\cite{Matlab2021}. 
The architectures selected for comparison are the baseline single channel (\textit{SC}, no form of FI detection present), the \textit{MA}~\cite{Mehmed2019}, the \textit{FWM}~\cite{Mehmed2020}, \textit{RSS}~\cite{weast2020sensors} and the Safety Shell architecture. The Safety Shell is tested in both 2 channel (\textit{SaS2}) and 3 channel (\textit{SaS3}) configurations.

The WMs used in all architectures have the same capabilities in every included channel.
All WMs will have accurate predictions of future motions of all observed objects, up to 3 seconds in the future, unless a FI affects the prediction quality.
This approach likely overestimates the capabilities of the safety channels' WMs, as they are normally simplified and consequently less capable (see Section \ref{section:existingPatterns}).
Safety channels include a WM and some form of fallback MP, while nominal AD channels (as also used as parallel channels in SaS2 and SaS3 architectures) include a WM and a nominal MP.
All nominal MPs use a sampled motion planner, derived from Matlab's example library \cite{Matlab2021}.

To increase the realism of the tested comparison, we assume that a third channel in SaS3 will have a limited capability MP.
Cost concerns of system design may lead to the implementations of progressively less desired and capable AD channels.
To simulate this, the 3rd AD channel is set up to continue the journey at a maximum speed of 50 km/h or the intended maximum speed, whichever is lower.

To evaluate the different architectures, {we compare} safety and availability.
Results are judged based on the percentage of collisions over the tested scenarios and the ability to reach the goal position in a scenario within reasonable time, respectively.
Longitudinal and lateral accelerations are tracked throughout the simulation, and used as a relative measure of comfort, when relevant.
Because the used MP trajectory planners and vehicle dynamics are rudimentary, only relative conclusions on comfort can be drawn from these simulations.

Table \ref{tab:overviewOfFNTests} shows the all tested FIs and the scenarios the respective FIs are tested with. 
The scenario selection is based on predominant severe crash scenarios according to \cite{safeup2021usecases,nitsche2017pre}.
We restrict testing to speeds ranging from 8 to 25 m/s (29 to 90 km/h).
As the majority of fatal collisions occur on non-highway roads and as the risk of fatal collisions below 29 km/h is limited~\cite{Richards2010}, this range allows for appropriate scenarios.
\begin{table*}[]
\centering
\caption{Descriptions of FIs that are tested in the numerical simulations}
\label{tab:overviewOfFNTests}
\resizebox{\textwidth}{!}{%
\begin{tabular}{@{}llll@{}}
\toprule
Test&Type of FI & Figure & Description of the effect on affected channels \\ \midrule
1 & Wrong predicted motion & Fig. \ref{fig:SR_Occlusionswerve_3lane} & The merging vehicle is predicted to slow and stop, instead of merging into the ego's path. \\
2 & Missed object & Fig. \ref{fig:SR_VRU_InLane} & The pedestrian walking in the ego lane is not detected. \\
3 & Missed object & Fig. \ref{fig:TJ_VRUAB_VehicleC}.C & The vehicle approaching the T-junction in the opposite lane is not detected. \\
4 & Incorrect object location & Fig. \ref{fig:TJ_VRUAB_VehicleC}.B & The pedestrian B approaching the T-junction is observed to be further away from the T-junction. \\
5 & Incorrect object location & Fig. \ref{fig:SR_Ego_AC_VRU_DB_crossingEgoLane} & The pedestrian crossing the ego lane is observed {to be far from} the ego lane. \\
6 & Dangerous trajectory & Fig. \ref{fig:SR_Occlusionswerve_3lane} & The trajectory does not appropriately deal with the predicted merging vehicle. \\
7 & Dangerous trajectory & Fig. \ref{fig:SR_VRU_InLane} & The trajectory does not appropriately deal with the observed pedestrian in lane. \\ 
8 & Simultaneous missed objects & Fig. \ref{fig:SR_VRU_InLane_Vehicle_Adjacent} & WM$_1$ does not detect the pedestrian while WM$_2$ (if present) does not detect the vehicle.\\
9 & Ghost object detection & Fig. \ref{fig:SR_VRU_InLane} & A pedestrian is detected ahead of the ego on its path, despite no pedestrian being present. \\
10 & Ghost object detection & Fig. \ref{fig:TJ_VRUAB_VehicleC}.A & A pedestrian is detected to cross the T-junction, despite no pedestrian being present. \\ 
\bottomrule
\end{tabular}
}%
\end{table*}

Tests 1 through 7 are run for both a single channel suffering an FI occurrence and for two channels simultaneously suffering the same FI occurrence.
These tests are run for target speeds of $8, 9, 10, ..., 25$ m/s.
The FWM architecture is excluded from test 1, as \cite{Mehmed2020} have not detailed how to deal with diverging predictions.

{To simulate a late obstacle detection}, tests 2 and 3 are repeated with the WMs of both channel 1 and 2 initially not observing the object, and the 2\textsuperscript{nd} channel identifying the object at a time of $0.1,0.2,0.3,..., 1.1$ s prior to collision.
The 3rd channel (if present in an architecture) is left unaffected and as such detects the obstacle immediately.
These forms of tests 2 and 3 are run at target speeds of $8, 17$ and $25$ m/s.

Test 8 is performed assuming that both channels never detect their respective missing object, yet, because each channels' WM misses a different object, all required information is present in a system with at least 2 channels to plot a safe path past the objects.
Tests 9 and 10 are performed with the second channel's WM (if present) detecting the ghost object for the full scenario duration and for a limited time, ranging from 2 s prior to expected impact to 0.5 s after expected impact.
Tests 8 through 10 are run for target speeds of $8, 9, 10, ..., 25$ m/s.

\begin{figure}[ht]
    \centering
    \includegraphics[width=6cm]{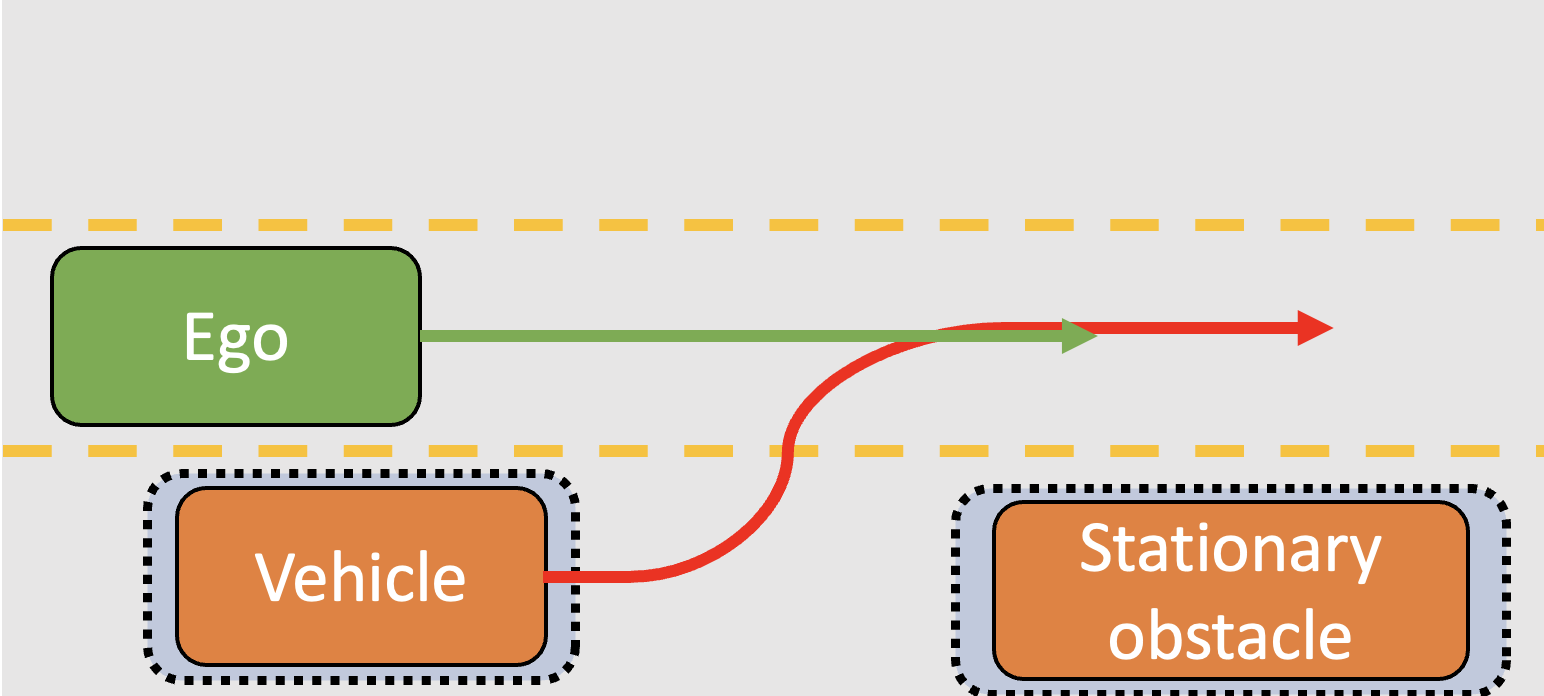}
    \caption{Three lane road scenario with a stationary obstacle forcing an adjacent vehicle to stop or swerve into our lane.}
    \label{fig:SR_Occlusionswerve_3lane}
\end{figure}
\begin{figure}[ht]
    \centering
    \includegraphics[width=6cm]{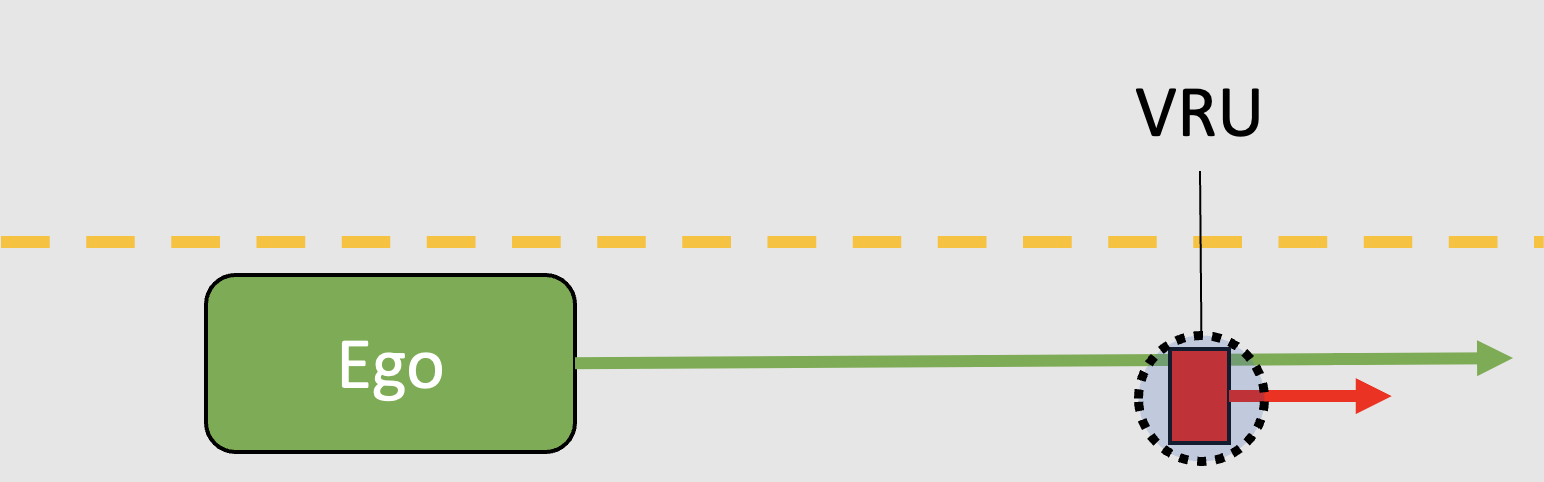}
    \caption{Straight road scenario, with a single pedestrian in the ego lane.}
    \label{fig:SR_VRU_InLane}
\end{figure}
\begin{figure}[ht]
    \centering
    \includegraphics[width=6cm]{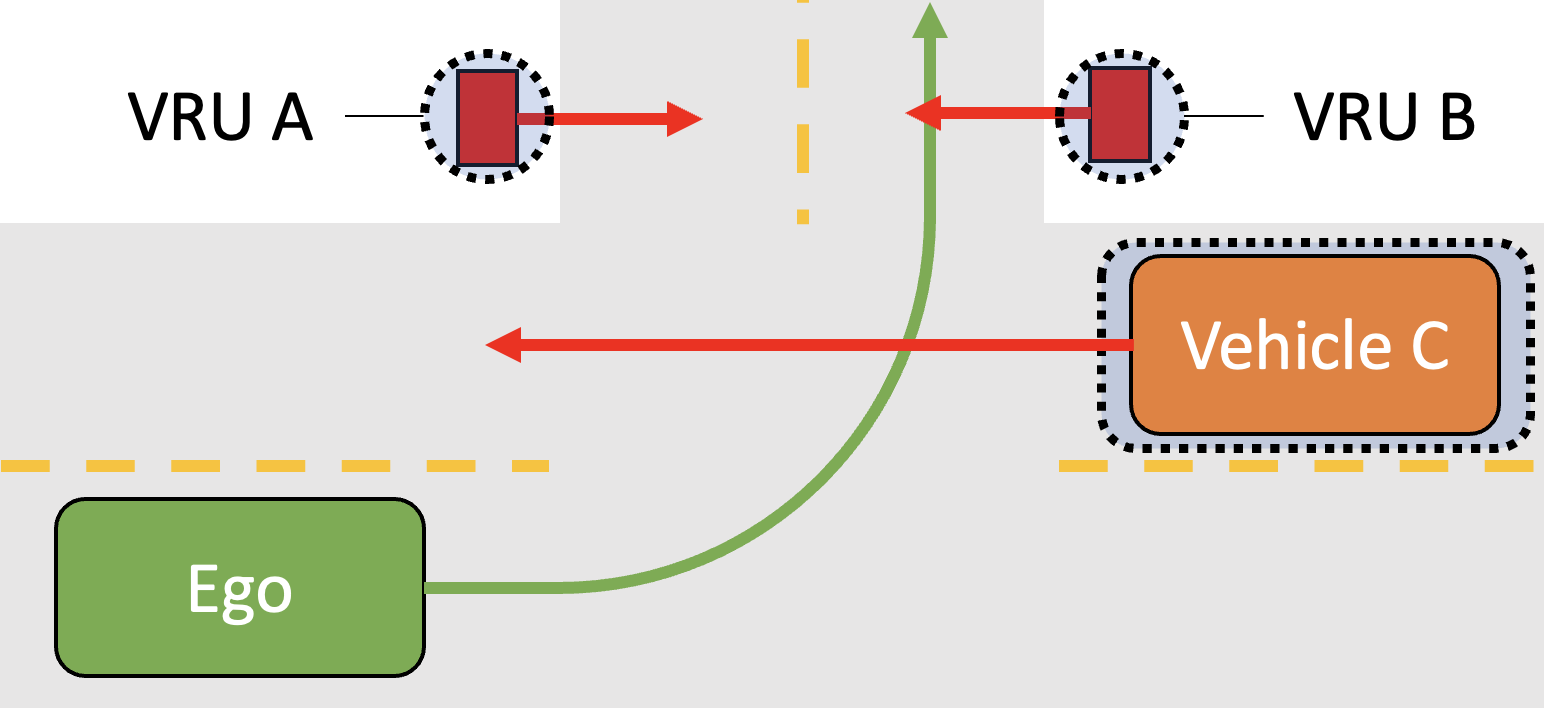}
    \caption{Left turn T-junction scenario, with three variations of the scenario shown: A. pedestrian crossing in the same direction as the ego vehicle, B. pedestrian crossing in opposite direction, C. Vehicle in the opposite direction lane.}
    \label{fig:TJ_VRUAB_VehicleC}
\end{figure}
\begin{figure}[h]
    \centering
    \includegraphics[width=6cm]{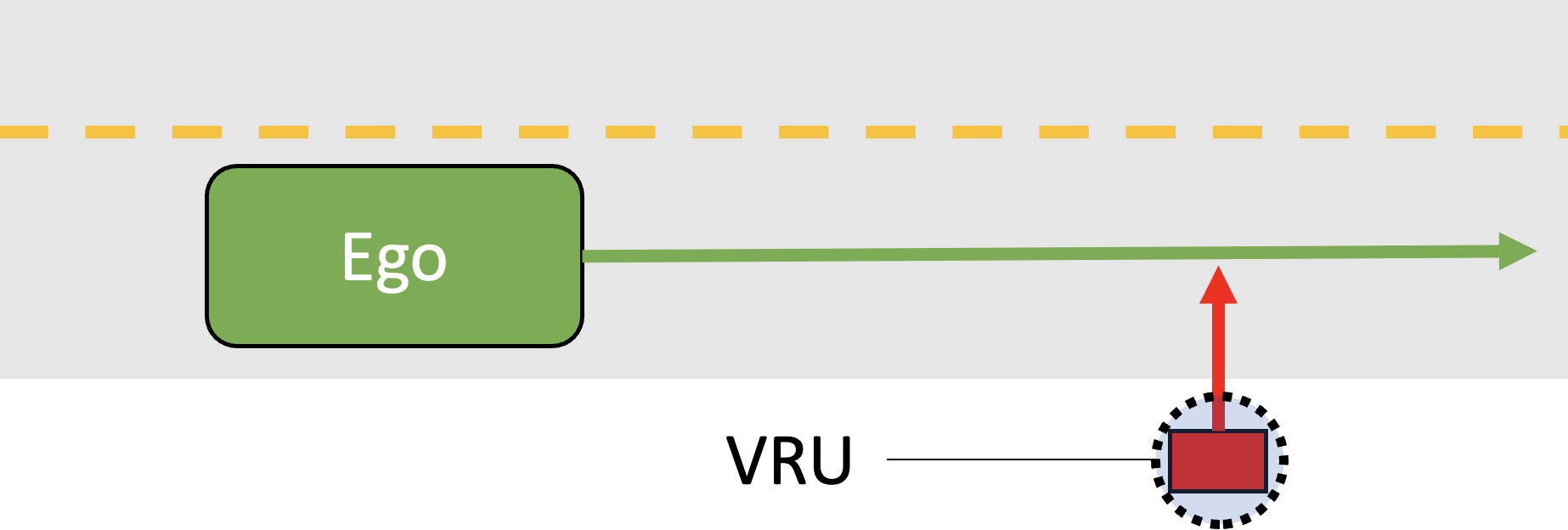}
    \caption{Straight road scenario, with a single pedestrian approaching the ego lane to cross it.}
    \label{fig:SR_Ego_AC_VRU_DB_crossingEgoLane}
\end{figure}
\begin{figure}[ht!]
    \centering
    \includegraphics[width=6cm]{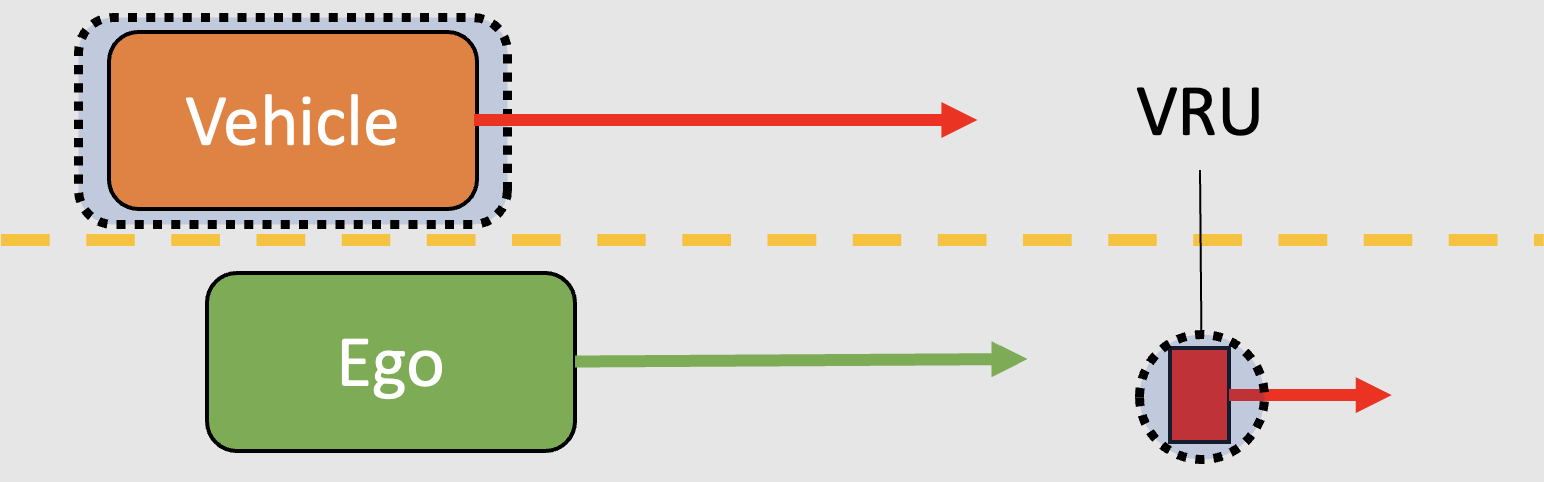}
    \caption{Straight road scenario, with a pedestrian in the ego lane and a vehicle with the same initial speed as the ego vehicle in the adjacent lane.}
    \label{fig:SR_VRU_InLane_Vehicle_Adjacent}
\end{figure}

\subsection{Safety Shell parameter settings}
The set of Safety Shell time parameters is shown in Table \ref{tab:SafetyShellSettings}. 
Note that $r_1$ is currently zero, as the short scenarios do not affect longer-term preference changes, as intended with \eqref{eq:ConsiderationTimeCalculation}.
The risk threshold is set to $R_{\mathrm{Threshold}}=0.25$.
{The selected level of $R_{\mathrm{Threshold}}$ is determined in conjunction with the tuning of the probability approximating safety metrics as listed in Appendix} \ref{appendix:risk} {and the severity metrics as listed in Appendix} \ref{appendix:severityScore}.
{It represents an approximation of the subjective feeling of unreasonable risk of the experimenter. 
Consequently, further optimizations are possible to refine both the threshold and the probability approximating metrics. }
\begin{table}[]
\centering
\caption{Used Safety Shell settings for numerical experiments}
\label{tab:SafetyShellSettings}
\begin{tabular}{@{}ll@{}}
\toprule
Parameter & Value \\
\hline
$\tau_{\mathrm{suff}}$ & 1.9 s \\
$\tau_{\mathrm{immediate}}$ & 0.4 s \\
$\tau_{\mathrm{C},1}^* \Delta_p$ & 1.8 s \\
$\tau_{\mathrm{C},2}^* \Delta_p$ & 1.5 s \\
$\tau_{\mathrm{C},3}^* \Delta_p$ & 1.0 s \\
$r_1$                   & 0 [1]\\
$q$                     & 20 [1]\\
$\Delta_s$              & 0.1 [s]\\
$\Delta_p$         & 0.1 [s]\\
\bottomrule
\end{tabular}
\end{table}

\subsection{Results}
\label{section:Results}
{We will first illustrate the Safety Shell's behaviour using an example from the tested cases. 
Next, we will compare the different architectures using the different variations of the scenarios.}

\subsubsection{Example of SaS2 behavior}
\begin{figure}[ht]
    \centering
    \includegraphics[width=8.4cm]{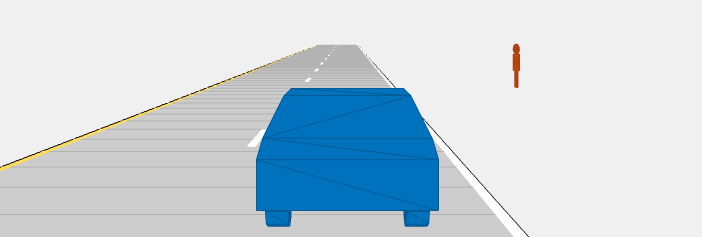}
    \includegraphics[width=8.4cm]{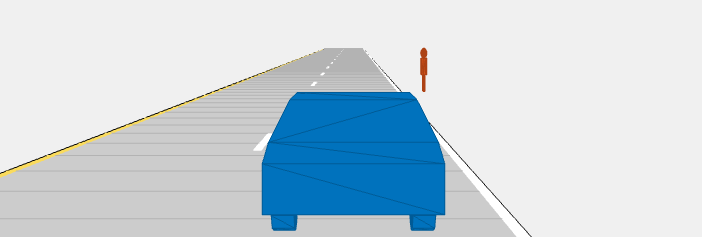}
    \vspace{-3pt}
    \caption{The two perspectives in test 5, with on top WM$_1$ and below WM$_2$. 
    WM$_1$ believes the pedestrian to be some distance away from the road, while WM$_2$ believes the pedestrian to be close to the road.}
    \label{fig:ChannelxViewLocationDiff}
    \vspace{-1pt}
\end{figure}

\begin{figure}[h]
    \centering
    \includegraphics[width=8.4cm]{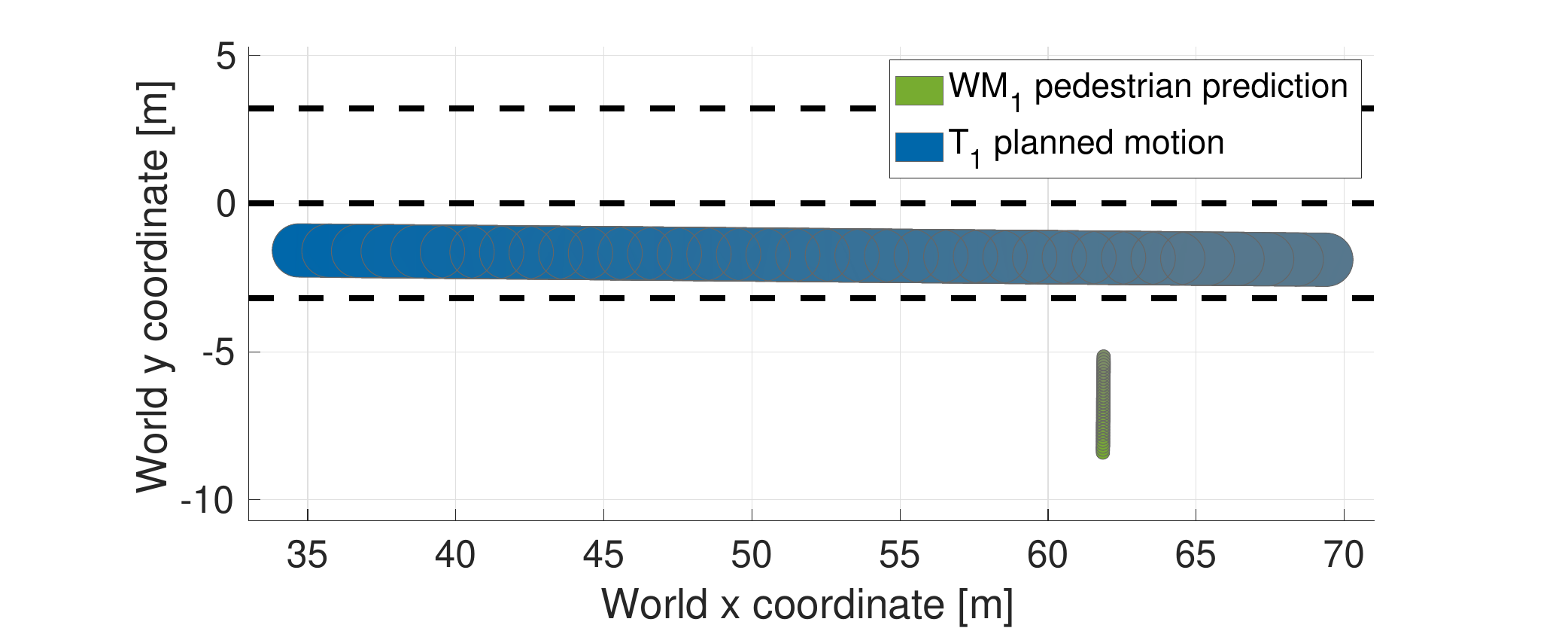}
    \includegraphics[width=8.4cm]{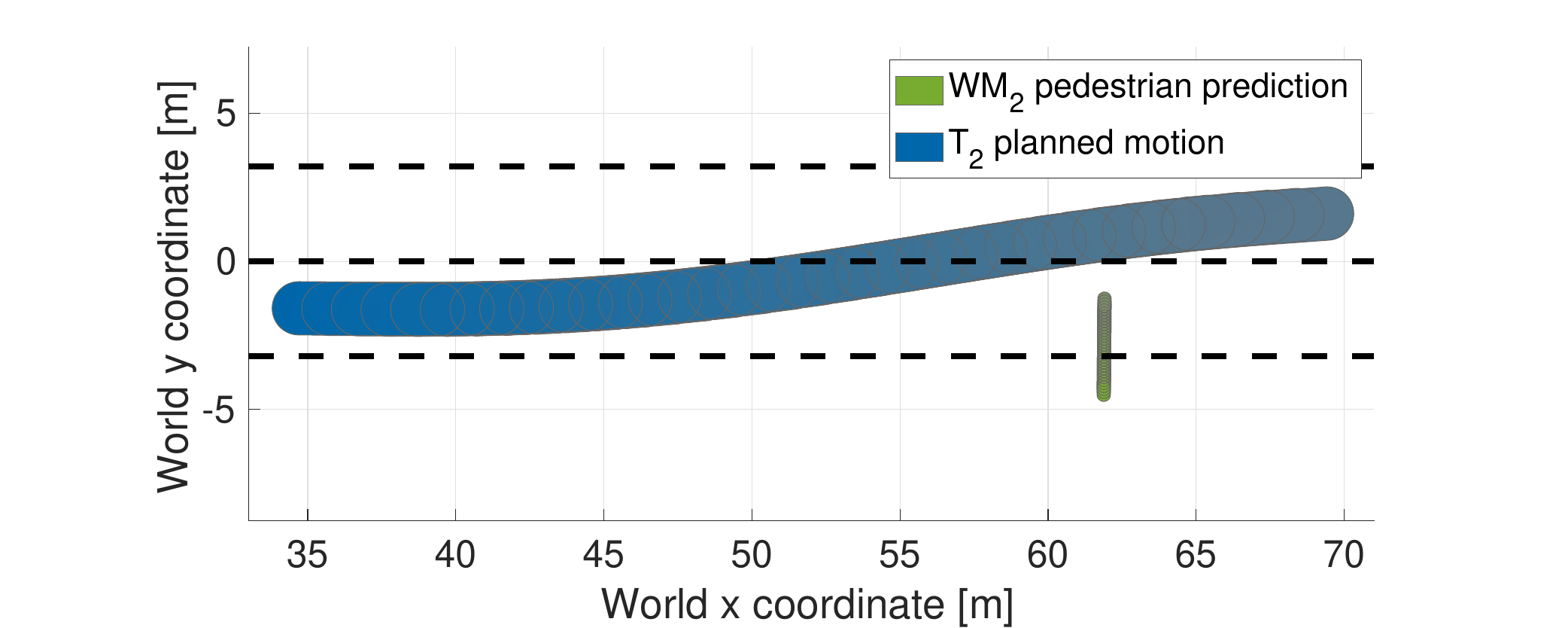}
    \includegraphics[width=8.4cm]{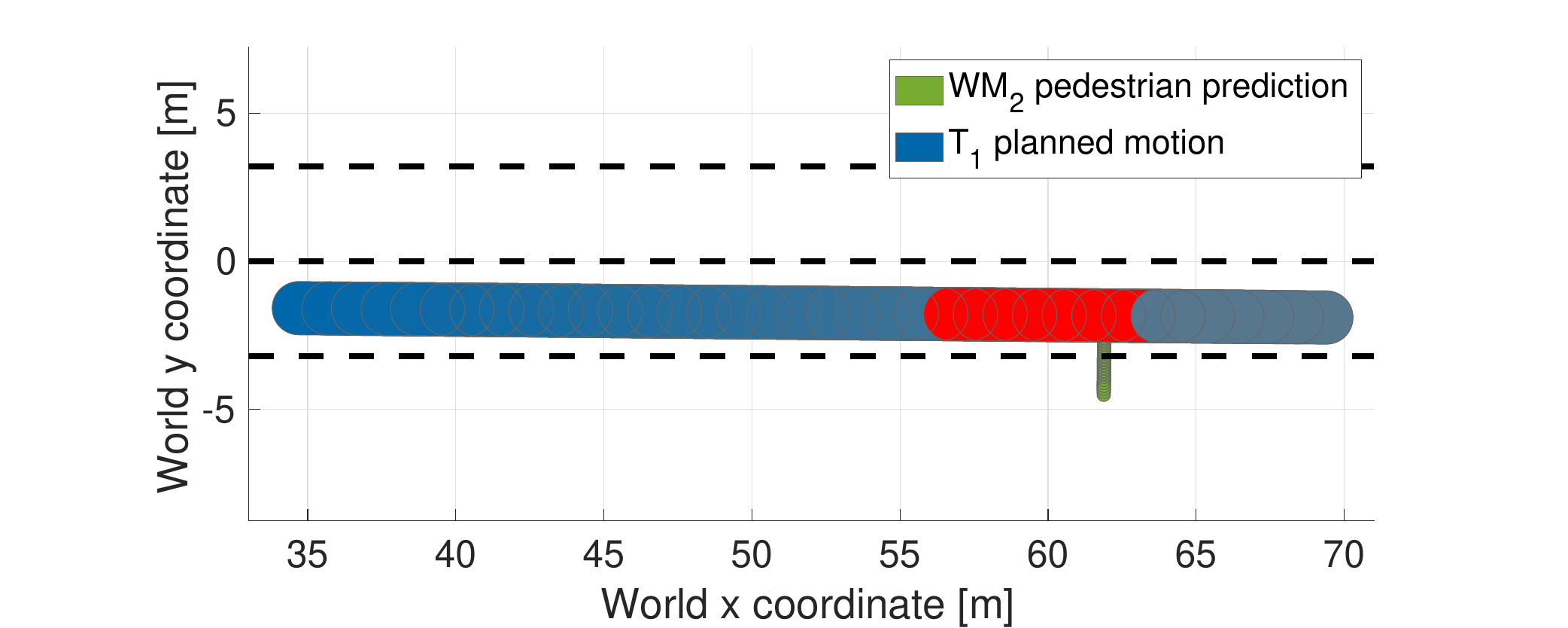}
    \caption{The channels' perception and prediction representation, with on top WM$_1$ showing the pedestrian a safe distance away from the road, the middle showing WM$_2$ observing the pedestrian dangerously close to the road and below the cross-check of channel 1 trajectory $T_1$ with channel 2 WM$_2$, with predicted collisions in red.}
    \label{fig:exampleWMXIncorrectLocationSaS2}
\end{figure}

\begin{figure}[ht]
    \centering
    \includegraphics[width=8.4cm]{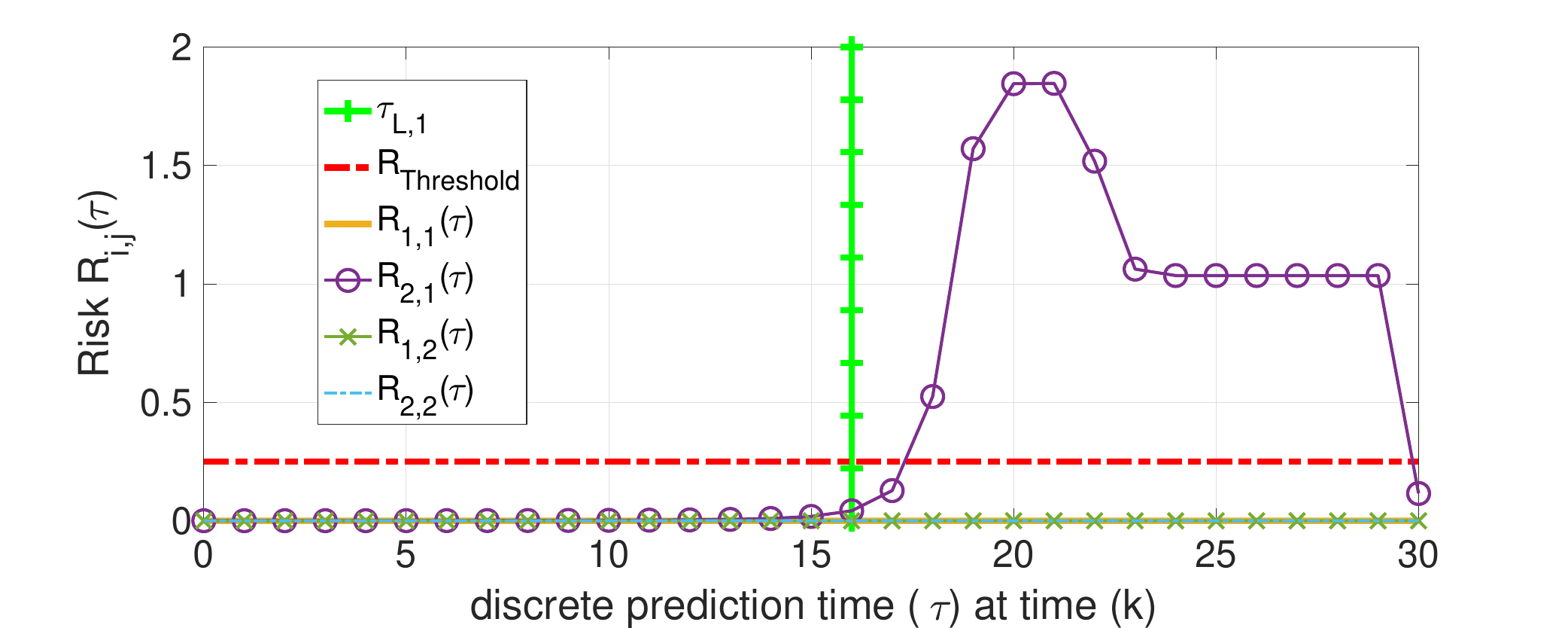}
    \includegraphics[width=8.4cm]{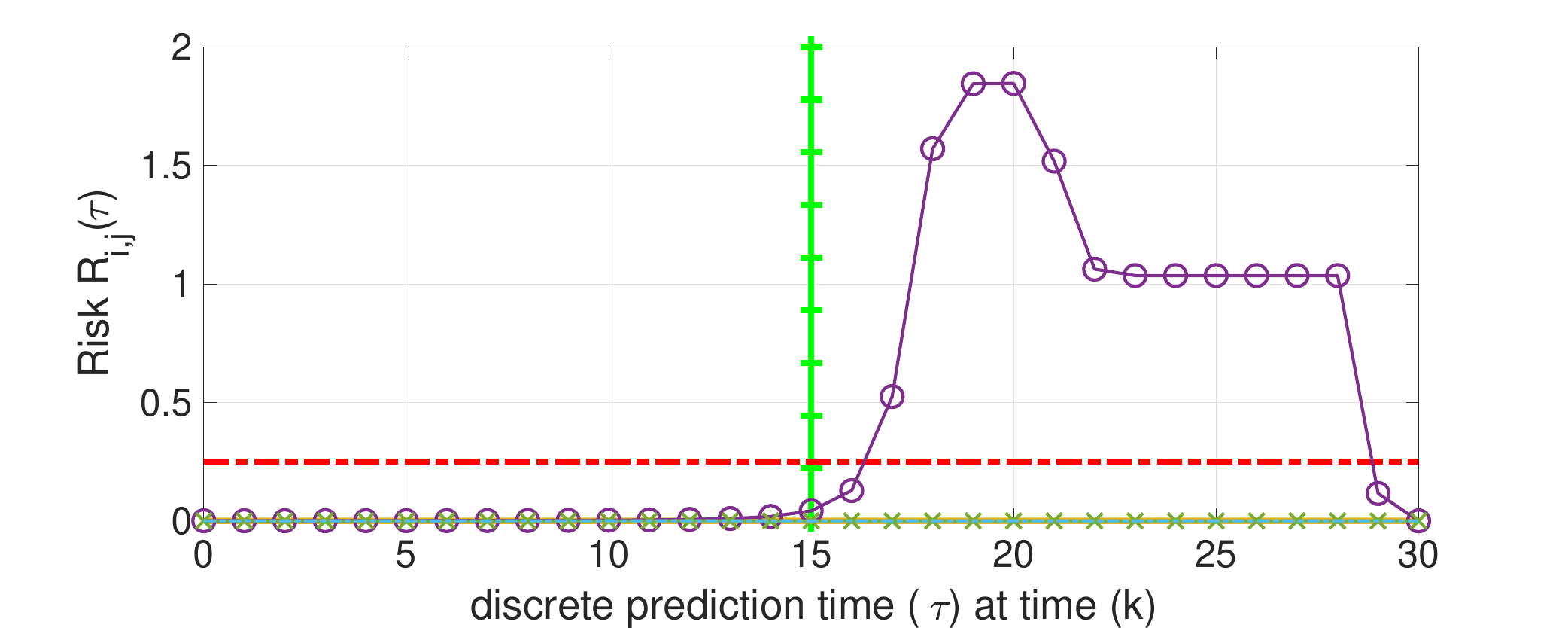}
    \caption{{Two graphs with the risk $R_{ij}(\tau)$ profiles over the predicted time horizon $\tau_{\mathrm{H}}$, resultant from the scenario in test 5. The top figure represents the situation 1 time step prior to switching, while the bottom graph represents the risk situation at the time of switching.}}
    \label{fig:exampleRiskFigureIncorrectLocationSaS2}
\end{figure}

{To illustrate the SaS2 architecture and arbitration approach, we highlight test 5 based on the layout shown in Fig. }\ref{fig:SR_Ego_AC_VRU_DB_crossingEgoLane}.
We select test 5 as the incorrect location determination of a pedestrian meaning to cross a road is both intuitively understandable, visually distinctive FI {and it mirrors the example provided in Section} \ref{section:highLevelSafetyShellLogic}{.}
Fig. \ref{fig:ChannelxViewLocationDiff} shows the difference in the perceived environment of the two WMs {at a time we will denote as $k$}, while Fig. \ref{fig:exampleWMXIncorrectLocationSaS2} shows the internal channel view of the scenario, including the predicted motions of the pedestrian and the planned trajectories. 
Because WM$_2$ observes the pedestrian close to the road and predicts the pedestrian to move onto the road, the cross-check of WM$_2$ with $T_1$ {predicts a collision if we continue along this trajectory and, consequently, }results in high risk values $R_{21}(\tau)$, as illustrated in the red overlay in the bottom of Fig. \ref{fig:exampleWMXIncorrectLocationSaS2}.
{The top graph in Fig.} \ref{fig:exampleRiskFigureIncorrectLocationSaS2} {shows the results of the cross-channel risk analysis, where the WM$_2$ evaluation of $T_1$ results in a $R_{21}(\tau)$ graph that exceeds the unreasonable threshold at $\tau_{\mathrm{U},1}(k)=$18.
The last safe intervention time for channel 1 is $\tau_{\mathrm{L}1}(k)=16$ or $\Delta_p\tau_{\mathrm{L}1}(k)=1.6$s.
Given that channel 1 is the more preferred channel and that} \eqref{eq:SetOfSafetySwitchChannels2} {rules out channel 2, as the preference value of channel 2 of $\tau_{\mathrm{C},2}(k)=15$ (see Table} \ref{tab:SafetyShellSettings}{) is smaller than }$\tau_{\mathrm{L}1}(k)$, {the system continues with channel 1 ($C(k) = 1$), providing channel 1 the opportunity to revise its motion plan to a safer alternative.}

{In test 5, given that channel 1 has a persistent FI in its WM, the channel 1 $T_1$ is still dangerous in the next time-step $k+1$, represented by the bottom graph in Fig.} \ref{fig:exampleRiskFigureIncorrectLocationSaS2}.
Through \eqref{eq:LSI_calc_generalized} the last safe intervention time $\tau_{\mathrm{L}1}(k+1)$ is computed as $15\Delta_p=1.5$s, which equals the $\tau_{\mathrm{C},2}(k+1)$ and, consequently, the SaS2 arbiter follows \eqref{eq:safetyDrivenPreferredChoice2} to switch to channel 2 and safely continues the journey. 

\subsubsection{Tests 1 through 7 with the first channel}
{Through the numerical results comparison in} Table \ref{tab:ResultsOverviewSingleChannel} the advantage of any type of FI detection is obvious, as all architectures except the single-channel architecture are able to identify when the preferred channel proposes a dangerous trajectory.
However, the different approaches of the tested architectures impact the resultant availability, as is shown in the lower part of Table \ref{tab:ResultsOverviewSingleChannel}.
The basic MA architecture is unable to continue and therefore fails to maintain availability, as its arbitration logic disengages the nominal channel and activates its fallback MP if a collision is predicted by the safety channel \cite{Mehmed2019}.
The FWM architecture is able to continue through the conservative free space fusion~\cite{Mehmed2020} (see Fig. \ref{fig:Mehmed2020}) 
in case of missed object detections and incorrect object location issues in the primary channel (tests 2 through 5).
However, this architecture is cannot continue if the primary channel's MP system generates an unsafe trajectory, as it must use its fallback MP and come to a stop (test 6 and 7).
The RSS architecture similarly fails to reach the goal position in tests 2 and 7, both based on the scenario of a pedestrian walking in the ego lane (Fig. \ref{fig:SR_VRU_InLane}).
As the RSS architecture does not include an extra motion planner (see Fig. \ref{fig:weast2020}), it is unable to plan a route around an object on path if the primary system fails to account for it.

\begin{figure}[ht!]
    \centering
    \includegraphics[width=8.4cm]{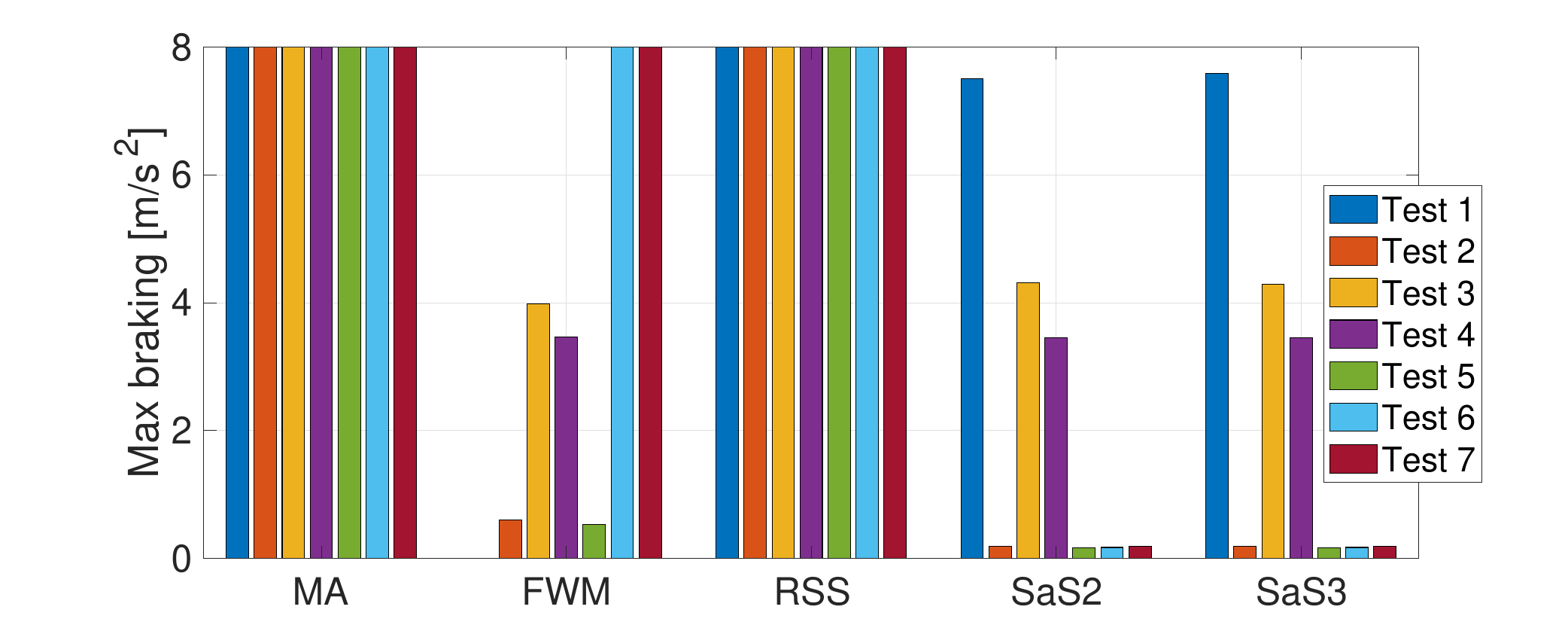}
    \caption{The peak braking deceleration of each test, averaged over all tested speeds, for a FI occurring only in the primary channel.}
    \label{fig:1ChannelFNBraking}
\end{figure}

Because of the necessary switch to either the fallback MP (MA and FWM) or the enforced motion restrictions (RSS), the comfort of passengers is severely impacted, as seen in Fig. \ref{fig:1ChannelFNBraking}.
By contrast, the SaS2 and SaS3 architectures are able to employ one of the parallel channels to continue without resorting to emergency braking in tests 2 through 7 and in test 1 for speeds less than 10 m/s. 
Throughout all tests reported in Table \ref{tab:ResultsOverviewSingleChannel}, the SaS3 architecture manages to attain the same speeds as the SaS2 architecture, despite the SaS3 architecture having a slower, safe third channel. 
This is due to the arbitration logic that selects the faster more preferred channel 2, through \eqref{eq:chooseMostPreferredChannel2}-\eqref{eq:escapeConsideration3} in all these cases.

\subsubsection{Tests 1 through 7 affecting the first and second channel}
\begin{table*}[]
\centering
\caption{Results of a FI occurring in the first channel for the entire simulation duration}
\label{tab:ResultsOverviewSingleChannel}
\begin{tabular}{cllllllll}
\toprule
\multicolumn{2}{l}{} & Test (Table \ref{tab:overviewOfFNTests}) & SC & MA & FWM & RSS & SaS2 & SaS3 \\ \hline
\multirow{4}{*}{Collisions} & Incorrect prediction & 1 & 78\% & 0\% & n.a. & 0\% & 0\% & 0\% \\
 & Missed object & 2 \& 3 & 100\% & 0\% & 0\% & 0\% & 0\% & 0\% \\
 & Incorrect obj. location & 4 \& 5 & 100\% & 0\% & 0\% & 0\% & 0\% & 0\% \\
 & dangerous trajectory & 6 \& 7 & 100\% & 0\% & 0\% & 0\% & 0\% & 0\% \\ \hline
\multirow{4}{*}{Availability} & Incorrect prediction & 1 & 22\% & 0\% & n.a. & 100\% & 100\% & 100\% \\
 & Missed object & 2 \& 3 & 0\% & 0\% & 100\% & 50\% & 100\% & 100\% \\
 & Incorrect obj. location & 4 \& 5 & 0\% & 0\% & 100\% & 100\% & 100\% & 100\% \\
 & dangerous trajectory& 6 \& 7 & 0\% & 0\% & 0\% & 50\% & 100\% & 100\% \\ \hline
\end{tabular}
\end{table*}

\begin{table*}[]
\centering
\caption{Results of an identical FI occurring in the first and second channel simultaneously for the entire simulation duration}
\label{tab:ResultsOverviewTwoChannel}
\begin{tabular}{cllllllll}
\toprule
\multicolumn{2}{l}{} & Test (Table \ref{tab:overviewOfFNTests}) & SC & MA & FWM & RSS & SaS2 & SaS3 \\ \hline
\multirow{4}{*}{Collisions} & Incorrect prediction & 1 & 78\% & 72\% & n.a. & 0\% & 72\% & 0\% \\
 & Missed object & 2 \& 3 & 100\% & 100\% & 100\% & 0\% & 100\% & 0\% \\
 & Incorrect obj. location & 4 \& 5 & 100\% & 100\% & 100\% & 0\% & 100\% & 0\% \\
 & dangerous trajectory & 6 \& 7 & 100\% & 0\% & 0\% & 0\% & 0\% & 0\% \\ \hline
\multirow{4}{*}{Availability} & Incorrect prediction & 1 & 22\% & 0\% & n.a. & 100\% & 28\% & 94\% \\
 & Missed object & 2 \& 3 & 0\% & 0\% & 0\% & 50\% & 0\% & 100\% \\
 & Incorrect obj. location & 4 \& 5 & 0\% & 0\% & 0\% & 100\% & 0\% & 100\% \\
 & dangerous trajectory & 6 \& 7 & 0\% & 0\% & 0\% & 50\% & 50\% & 100\% \\ \hline
\end{tabular}
\end{table*}

\begin{table*}[]
\centering
\caption{Availability and the peak braking averaged over the evaluated speeds of each false-positive test. }
\label{tab:FPResults}
\resizebox{\textwidth}{!}{%
\begin{tabular}{@{}clllllll@{}}
\toprule
\multicolumn{2}{l}{} & Test (Table \ref{tab:overviewOfFNTests}) & MA & FWM & RSS & SaS2 & SaS3 \\ 
\hline
\multirow{2}{*}{Availability} & Ghost pedestrian in lane & 9 & 0\% & 100\% & 0\% & 100\% & 100\% \\
 & Ghost pedestrian in left turn & 10 & 0\% & 100\% & 89\% & 100\% & 100\% \\ 
 \hline
\multirow{2}{*}{Braking deceleration} & Incorrect prediction & 9 & 8 m/s$^{2}$ & 0.6 m/s$^{2}$ & 8 m/s$^{2}$ & 0.2 m/s$^{2}$ & 0.2 m/s$^{2}$ \\
 & Missed object & 10 & 8 m/s$^{2}$ & 3.5 m/s$^{2}$ & 8 m/s$^{2}$ & 3.6 m/s$^{2}$ & 3.6 m/s$^{2}$ \\ \bottomrule
\end{tabular}
}%
\end{table*}
In Table \ref{tab:ResultsOverviewTwoChannel} the increase in collisions across MA, FWM and SaS2 architectures is evident.
Because the WMs of both channels present in those architectures suffer from a simultaneous FI in tests 1 through 5, the architectures are unable to detect the danger.
In tests 6 and 7 those same architectures are still able to ensure safety.
The SaS2 architecture is able to continue in test 6, as the fallback MP brake manoeuvre creates sufficient distance between the merging vehicle and the ego vehicle (see Fig. \ref{fig:SR_Occlusionswerve_3lane}) and allows for the safe continued motion of the primary functionality.
A similar approach can be created for MA and FWM architectures, but was not explicitly included in their design~\cite{Mehmed2019,Mehmed2020}, so is not included here.

The availability figures in Table \ref{tab:ResultsOverviewTwoChannel} show the clear benefit of a the Safety Shell fitted with a third channel, despite this channels limitations. 
These limitations cause it to fail to reach the goal position in the 90 km/h case in test 1, as the reasonable duration to complete the scenario is also proportionally decreased based on the target speed. 
In contrast, the RSS system fails to maintain availability in all tests where an object remains in front of the vehicle (tests 2 and 7), highlighting again the relative inflexibility of envelope reductions to ensure safety.
Concerning the comfort of the intervention, the RSS interventions necessitate braking at 8 m/s$^2$, while the SaS3 architecture only necessitates this level of braking in test 1 runs of more than 10 m/s.
{All} other tested architectures either crash or execute maximum braking.

\subsubsection{Test 2 and 3, with delayed detection in channel 2}
\begin{figure}[ht]
    \centering
    \includegraphics[width=8.4cm]{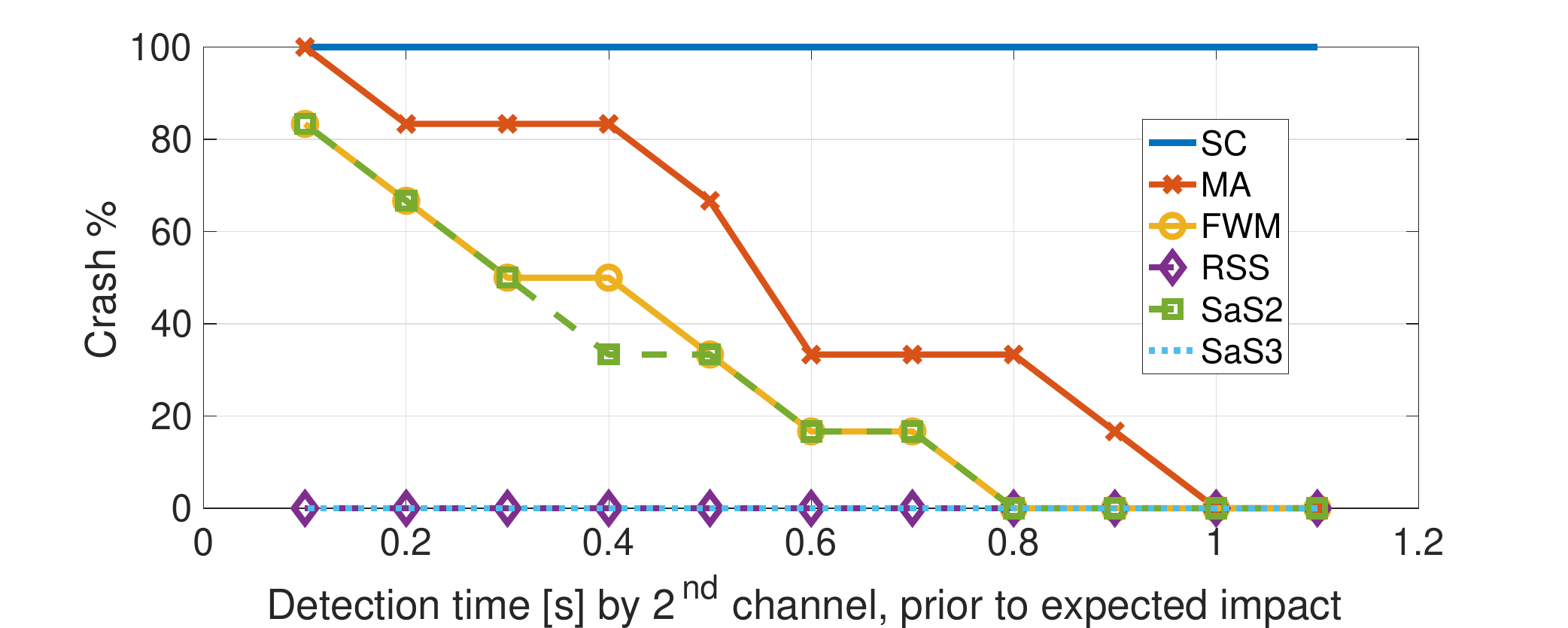}
    \caption{The percentage of tests that result in crashes, as a function of the detection time of the object by the 2\textsuperscript{nd} channel prior to expected impact.}
    \label{fig:2ChannelCrashTemp}
\end{figure}

The MA, FWM and SaS2 architectures improve collision avoidance performance as the detection time prior to expected impact is increased, as seen in Fig. \ref{fig:2ChannelCrashTemp}.
The SaS2 architecture shows improved performance compared to the MA architecture at all detection times.
This is because the observation of the object is only present in the second WM, and the currently implemented fallback MP uses an AEB that does not use evasive steering.
The SaS2 however uses \eqref{eq:escapeConsideration2} and \eqref{eq:escapeConsideration3} to be able to switch to an AEB along the least dangerous path available, in this case one that tries to evade the observed object. 
This brings the SaS2 architecture roughly on par with the FWM architecture in terms of Safety.
The sharp drop in collisions in the tested scenario at a detection very close to expected impact is due to the nature of scenario 1, where the pedestrian has only a very limited overlap to the nominal car trajectory.
Therefore, this scenario benefits from evasive steering for late detections.
Only the RSS and the Safety Shell with 3 channels are able to ensure safety in all tested scenarios, as both architectures have a third channel unaffected by the FI, that, consequently, correctly detects the objects missed by other channels. 
\begin{figure}[ht]
    \centering
    \includegraphics[width=8.4cm]{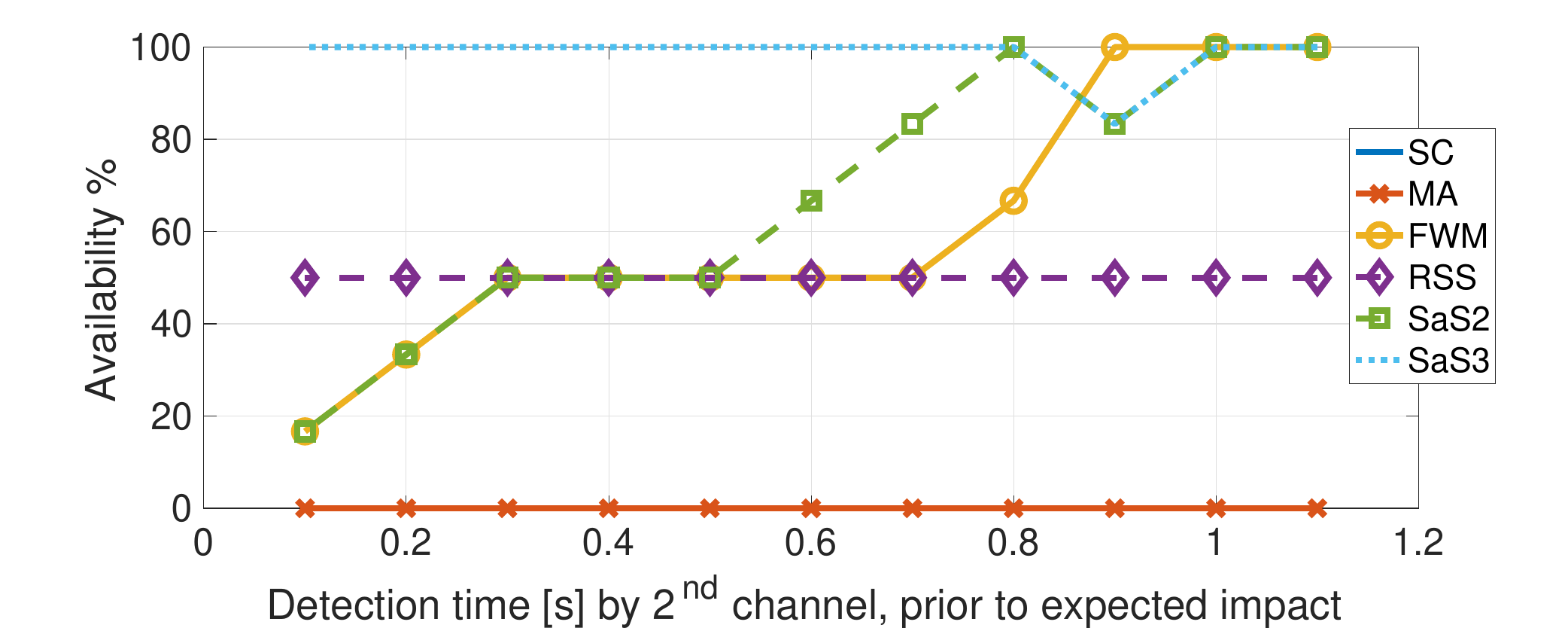}
    \caption{Availability of mission continuing capability as a function of the detection of the conflicting object prior to expected impact by the 2\textsuperscript{nd} channel.}
    \label{fig:2ChannelAvailabilityTemp}
\end{figure}

The MA and FWM architectures have reduced availability compared to the SaS2 architecture, as seen in Fig. \ref{fig:2ChannelAvailabilityTemp}. 
This is again due to the previously mentioned disengagement logic for the MA and FWM architectures.
The RSS architecture suffers poor availability compared to the equally safe SaS3 architecture due to the lack of flexible alternative MPs in test 2, as similarly observed in the prior results.

\subsubsection{Test 8, affecting 2 channels in different ways}
\begin{figure}[ht]
    \centering
    \includegraphics[width=8.4cm]{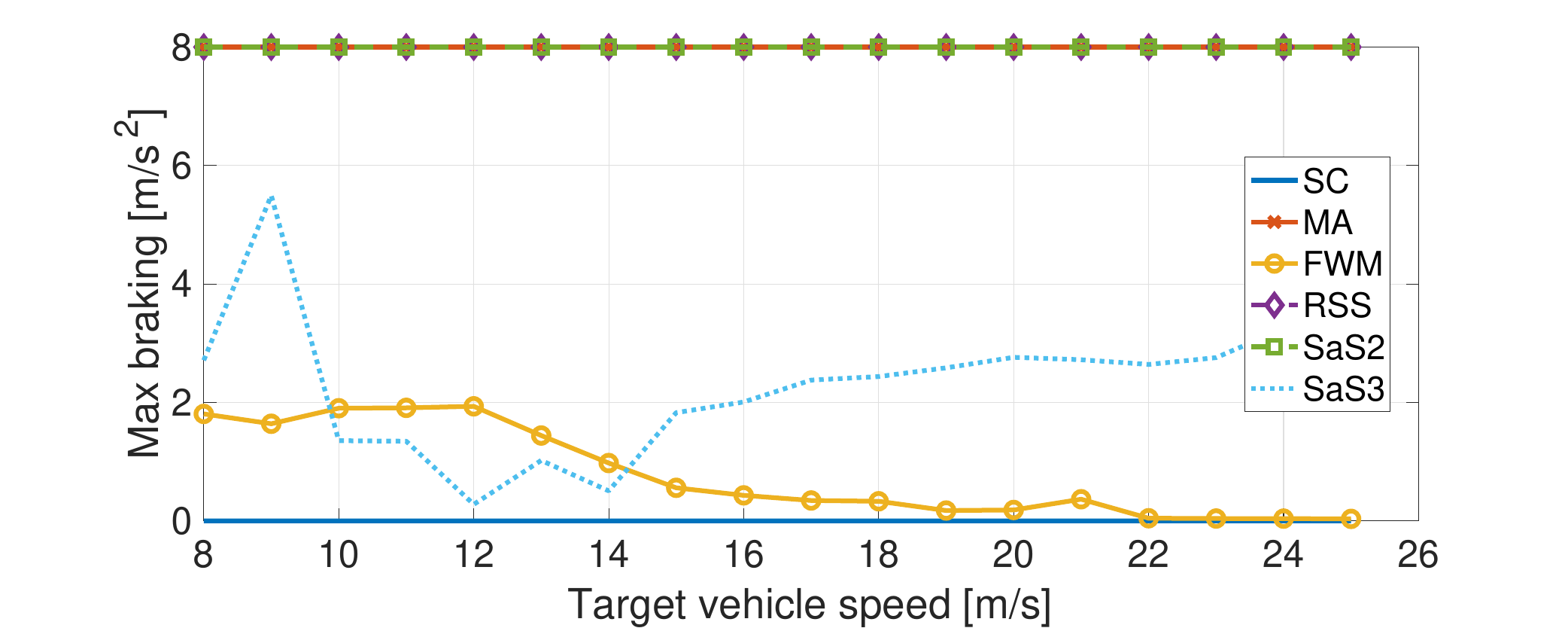}
    \caption{Maximum braking deceleration in case of simultaneous missed object detections of different objects for channel 1 and 2.}
    \label{fig:2ChannelBrakingDiffObj}
\end{figure}

Because WM$_1$ does not observe the pedestrian on its path, channel 1 plans a dangerous trajectory (see Fig. \ref{fig:SR_VRU_InLane_Vehicle_Adjacent}).
All architectures, except the SC architecture, see the danger in this trajectory and ensure safety.
The MA, RSS and SaS2 architectures do so via triggering their escape trajectories and, consequently, suffer poor comfort.
The FWM architecture uses its fused free space to generate a {relatively} comfortable trajectory past the observed objects.
This is shown through {reduced} peak braking values in Fig. \ref{fig:2ChannelBrakingDiffObj}.
The SaS3 architecture closely trails trails the FWM architecture in comfort.
We attribute the high peak braking at low speeds in the SaS3 architecture to the poorly tuned motion planner used in these experiments.
The SaS2 architecture is able to select its the second channels nominal MP to overtake the pedestrian and reach the goal position after slowing down sufficiently and letting the adjacent vehicle pass.
This shows that through the included escape trajectory planner, the 2-channel Safety Shell system can also remain safe in these complex mutually-excluding situations.

\subsubsection{{False Positive} Tests 9 and 10, affecting the second channel}
Table \ref{tab:FPResults} shows both the availability and effective intervention impact on comfort through the peak braking values, averaged over the tested speeds. 
The SC architecture is not included in this overview, as only the second channel suffers from the FI that causes it to see a ghost object. 
Consequently, the SC architecture remains unaffected.
The MA and RSS architectures both suffer from a reduced availability and require severe braking. 
The severe braking is due to both architectures enforcing a fixed intervention.
The reduction in availability is again due to the MA arbitration logic of disengaging upon intervention and the lack of flexible MP alternatives in the RSS architecture, respectively.
The FWM, SaS2 and SaS3 architectures all perform similarly. 
The relatively high braking in the test 10 can be attributed to the poor MP optimization, as the (unaffected) SC also brakes with a similar magnitude to be able make the turn in test 10, shown in Fig. \ref{fig:TJ_VRUAB_VehicleC}.A.

\section{Discussion and outlook}
\label{section:discussion}
A key first benefit of the Safety Shell architecture is its ability to maintain safety and availability of journey-continuing autonomous functionality. 
Indeed, Table \ref{tab:ResultsOverviewSingleChannel} shows that the Safety Shell architecture with two channels exceeds the availability of all other two-channel architectures in the face of hazardous scenarios.
Table \ref{tab:ResultsOverviewTwoChannel} shows that the three channel Safety Shell version exceeds the safety of all two-channel systems and outperforms all other architectures with respect to availability.
Table \ref{tab:FPResults} shows that both Safety Shell architectures return a 100\% availability score when faced with the evaluated false-positive test.
The challenging late object detection tests show an improvement in safety for the two-channel Safety Shell architecture when compared to the MA architecture in Fig. \ref{fig:2ChannelCrashTemp}, while excelling in overall availability as seen in Fig. \ref{fig:2ChannelAvailabilityTemp}. 

{The increased complexity of test 8 shows the ability of the Safety Shell to deal with complex scenarios, where channels mutually determine each other to be unsafe. 
Crowded scenarios with many relevant objects (e.g., partially blocked intersections, traversing through busy partially pedestrianized zones) may cause the cross-comparison to label many trajectories as violating the risk threshold at some point in the future.}
{The Safety Shell is intended to use availability maintaining and mission continuing channel choices through} \eqref{eq:SetOfMorePreferredSafeChannels2} {-} \eqref{eq:escapeConsideration2} {and, as a consequence, switch to the fallback escape trajectory as late and as little as possible.}
{Even when the situation arises that escape trajectory activation is required to maintain the safety of the vehicle, like other evaluated architectures listed in Table} \ref{table:overviewOfArchitectures}, {the Safety Shell can always switch back to a nominal channel the moment it detects that a channel has created a sufficiently safe trajectory, following} \eqref{eq:SetOfSafetySwitchChannels2}{ and rule 2.}
{We believe that the inclusion of multiple flexible nominal motion planning functions increases the chance that one of them is able to devise a way around the perceived risk causes, thereby preventing the Safety Shell from getting locked into the fallback trajectory.}
{In some cases, no motion of the vehicle may be sufficiently safe, rendering the vehicle immobile.}
{In these cases a measure of belief in the presence of objects may be weighted in the calculation of risk through an adjustment of} \eqref{eq:risk_simplified}. 
{Such a belief in the objects reported by a channel can be influenced either through automatic comparison between channels, through remote operator evaluation if the vehicle is at standstill, or a combination of both.
The study of these situations is considered future work.}

{The Safety Shell logic is dependent on a sufficiently capable fallback MP.
However, in the case of unavoidable unreasonable risk (i.e., each channel and the fallback trajectory encounters near immediate unreasonable risk), the arbitration system does not currently minimize the total exposure to risk.
In other words, at that point it does not allow for the selection of a channel if that exposes the vehicle or traffic participants to less risk than the fallback trajectory.
Future work may improve this logic to ensure a sufficient minimization of risk that incorporates a suitable ethical framework for these cases.}

The Safety Shell architecture allows automated vehicle system developers to pick and choose from the capable AD channels that are being developed, to combine them to form a sufficiently safe and capable vehicle.
As \cite{Fu2023} shows that functional insufficiencies remain the dominant reported cause for test disengagement, we propose that combining a number of AD channels may allow the elevation of otherwise highly capable systems in need of supervision (SAE L2) to a single combined highly automated unsupervised system (SAE L3 or higher).
{However, to the best of our knowledge there currently is no method to assess the level of heterogeneity of channels, nor what the level of overlap in the unknown FIs is likely to be.
Combined with the fact that FIs remain in AD channels after initial development (see Section} \ref{section:FiHandlingDuringDevelopment}{), this means that the Safety Shell architecture does not guarantee completeness of FI handling.
The level in which the Safety Shell increases FI handling coverage or the optimal number of channels to include is a topic of future research.}

Next, the role of the Safety Shell on fault and failure detection and mitigation is not explicitly addressed in this paper so far. 
We propose that any dangerous failure should in turn lead to the exposure of dangerous behaviour. 
The detection of electro-mechanical failures (e.g., brake failures, steering actuator failures) is out of the scope of the Safety Shell capability.
However, failures affecting the performance of individual AD channels, e.g., bit-flips causing a dangerous trajectory or a power failure to an AD channel computer, are detectable by the Safety Shell cross-checking algorithm. 
The Safety Shell is not a substitute for the application of rigorous functional safety principles, but is an addition for unforeseen failure modes that affect high-level automation functions.
Similarly, the Safety Shell AD channel performance tracking function (see \eqref{eq:ConsiderationTimeCalculation}) can be re-used to provide early-warning indicators of insufficient safety performance of an AD channel~\cite{ul4600,koopman2022continuous}, providing the developers of AD channels with warnings of new FIs in their creations while the other parallel AD channels maintain safety. 
We expect continuous monitoring functions of some form to become obligatory in highly automated vehicles and re-using the safety improvements provided by the Safety Shell can be a convenient application of this concept. 

Our framework allows us to compare safety and availability performance of highly automated driving architectures, and we believe it is the first numerical architecture comparison to date. 
However, the numerical study presented here, despite the 10 different tested scenarios, the range of speeds and types of simulated functional insufficiencies, is still limited compared to the infinite variation in traffic experienced during real driving. 
To generalize the conclusions of this study to all possible driving scenarios would be a fallacy. 
Therefore we aim to create a more realistic simulation environment in the near future, to allow us to test the value of the Safety Shell architecture when exposed to a much larger variation of scenarios and emergent functional insufficiencies in the used AD channels. 
Finally, we hope to investigate the application of the work of \cite{qiu2021reliability} to architectural safety and availability considerations, to investigate if more generalizable safety and availability conclusions are possible.

\section{Conclusion}
\label{section:conclusion}
In this paper we have introduced a multi-channel architecture design for an automated vehicle, referred to as the Safety Shell, to handle the inevitable remaining unknown functional insufficiencies.
As expected based on the underlying rationale of the Safety Shell, numerical simulations show that the Safety Shell is able to attain safety on par or better when compared to the current state of art of FI-handling architectures.
It attains these results through its novel online channel selection arbitration algorithm, which uses the Last Safe Intervention Time combined with a (time-varying) preference order of available AD channels.
Finally, the benefit of a third channel to test the safety of planned motion is evident through the tests with simultaneously occurring FIs and highlights the versatility of the Safety Shell. 
Future work will focus on testing this architecture in more realistic simulation environments to assess the likelihood of deadlock-like situations and assess needed developments regarding online consideration time updates or false positive mitigations.

\vspace{-2 pt}
\section*{Acknowledgments}
{We thank Yuting Fu, Jochen Seemann, and Tim Beurskens from NXP Semiconductors for fruitful discussions. This work was funded by the project NEON, through the Dutch Research Council (NWO) Crossover Programme (project number 17628).}

\bibliographystyle{abbrv-doi-narrow}
\bibliography{library} 

\begin{thebibliography}{10}
\renewcommand*{\sfdefault}{PTSansNarrow-TLF}

\bibitem{Armoush2010}
A.~Armoush.
\newblock {\em Design patterns for safety-critical embedded systems.}
\newblock PhD thesis, RWTH Aachen University, 2010.

\bibitem{safeup2021usecases}
A.~Bálint, V.~Labenski, M.~Köbe, C.~Vogl, J.~Stoll, L.~Schories, L.~Amann,
  G.~B. Sudhakaran, P.~H. Leyva, T.~Pallacci, M.~Östling, D.~Schmidt, and
  R.~Schindler.
\newblock D2.6 use case definitions andinitial safety-critical scenarios.
\newblock Technical report, SAFE-UP, 2021.

\bibitem{cheng2022logically}
C.-H. Cheng, T.~Schuster, and S.~Burton.
\newblock Logically sound arguments for the effectiveness of ml safety
  measures.
\newblock In {\em International Conference on Computer Safety, Reliability, and
  Security}, pp. 343--350. Springer, 2022.

\bibitem{Damerow2015risk}
F.~Damerow and J.~Eggert.
\newblock Risk-aversive behavior planning under multiple situations with
  uncertainty.
\newblock In {\em 18th International Conference on Intelligent Transportation
  Systems}, pp. 656--663. IEEE, 2015.

\bibitem{Decastro2018}
J.~DeCastro, L.~Liebenwein, C.~Vasile, R.~Tedrake, S.~Karaman, and D.~Rus.
\newblock Counterexample-guided safety contracts for autonomous driving.
\newblock In {\em International Workshop on the Algorithmic Foundations of
  Robotics}, pp. 939--955. Springer, 2018.

\bibitem{DMV2021}
{Department of Motor Vehicles}.
\newblock 2021 autonomous vehicle disengagement reports.
\newblock Accessed online June 13th, 2022, 2021.

\bibitem{Eggert2014risk}
J.~Eggert.
\newblock Predictive risk estimation for intelligent adas functions.
\newblock In {\em 17th International IEEE Conference on Intelligent
  Transportation Systems}, pp. 711--718. IEEE, 2014.

\bibitem{ferrari2022criteria}
E.~Ferrari, R.~Schlick, J.~L. De~la Vara, P.~Folkesson, and B.~Sangchoolie.
\newblock Criteria for the analysis of gaps and limitations of v\&v methods for
  safety-and security-critical systems.
\newblock In {\em International Conference on Computer Safety, Reliability, and
  Security}, pp. 35--46. Springer, 2022.

\bibitem{Fruehling2019}
T.~Fruehling, A.~Hailemichael, C.~Graves, J.~Riehl, E.~Nutt, R.~Fischer, and
  A.~K. Saberi.
\newblock Architectural safety perspectives \& considerations regarding the
  ai-based av domain controller.
\newblock In {\em International Conference on Connected Vehicles and Expo}, pp.
  1--10. IEEE, 2019.

\bibitem{Fu2023}
Y.~Fu, J.~Seemann, C.~Hanselaar, T.~Beurskens, A.~Terechko, E.~Silvas, and
  W.~P. M.~H. Heemels.
\newblock Characterization and mitigation of insufficiencies in automated
  driving systems.
\newblock In {\em The 27th International Technical Conference on the Enhanced
  Safety of Vehicles}, 2023.

\bibitem{Fu2020}
Y.~Fu, A.~Terechko, J.~F. Groote, and A.~K. Saberi.
\newblock A formally verified fail-operational safety concept for automated
  driving.
\newblock 2020.

\bibitem{Furst2018}
S.~F\"urst.
\newblock Scalable architecture for autonomous driving.
\newblock {\em 9th Vector Congress}, 2018.

\bibitem{gerchinovitz2022object}
S.~F. Gerchinovitz and B.~de~Grancey.
\newblock Object detection with probabilistic guarantees: A conformal
  prediction approach.
\newblock In {\em Computer Safety, Reliability, and Security. SAFECOMP 2022
  Workshops: DECSoS, DepDevOps, SASSUR, SENSEI, USDAI, and WAISE Munich,
  Germany, September 6--9, 2022, Proceedings}, vol. 13415, p. 316. Springer
  Nature, 2022.

\bibitem{Hanselaar2022}
C.~Hanselaar, E.~Silvas, A.~Terechko, and W.~Heemels.
\newblock Detection and mitigation of functional insufficiencies in autonomous
  vehicles: The safety shell.
\newblock {\em Intelligent Transporation Systems}, 2022.

\bibitem{Hogeveen2021}
P.~Hogeveen, M.~Steinbuch, G.~Verbong, and A.~Hoekstra.
\newblock The energy consumption of passenger vehicles in a transformed
  mobility system with autonomous, shared and fit-for-purpose electric vehicles
  in the netherlands.
\newblock {\em The Open Transportation Journal}, 15(1), 2021.

\bibitem{Ishigooka2018}
T.~Ishigooka, S.~Honda, and H.~Takada.
\newblock Cost-effective redundancy approach for fail-operational autonomous
  driving system.
\newblock In {\em 2018 IEEE 21st International Symposium on Real-Time
  Distributed Computing (ISORC)}, pp. 107--115. IEEE, 2018.

\bibitem{iso21448}
{Road vehicles -- Safety of the intended functionality (ISO Standard No.
  21448:2022)}.

\bibitem{iso26262}
{Road vehicles -- Functional safety (ISO Standard No. 26262:2018)}.

\bibitem{Jackson2019}
D.~Jackson, J.~DeCastro, S.~Kong, D.~Koutentakis, A.~Leong, A.~Solar-Lezama,
  M.~Wang, and X.~Zhang.
\newblock Certified control for self-driving cars.
\newblock In {\em 4th Workshop On The Design And Analysis Of Robust Systems},
  2019.

\bibitem{koopman2022continuous}
R.~Johansson and P.~Koopman.
\newblock Continuous learning approach to safety engineering.
\newblock In {\em Critical Automotive applications: Robustness \& Safety},
  2022.

\bibitem{kirovskii2019driver}
O.~Kirovskii and V.~Gorelov.
\newblock Driver assistance systems: analysis, tests and the safety case. iso
  26262 and iso pas 21448.
\newblock In {\em IOP Conference Series: Materials Science and Engineering},
  vol. 534, p. 012019. IOP Publishing, 2019.

\bibitem{kochanthara2021functional}
S.~Kochanthara, N.~Rood, A.~K. Saberi, L.~Cleophas, Y.~Dajsuren, and M.~van~den
  Brand.
\newblock A functional safety assessment method for cooperative automotive
  architecture.
\newblock {\em Journal of Systems and Software}, 179:110991, 2021.

\bibitem{konighofer2020shield}
B.~K{\"o}nighofer, F.~Lorber, N.~Jansen, and R.~Bloem.
\newblock Shield synthesis for reinforcement learning.
\newblock In {\em International symposium on leveraging applications of formal
  methods}, pp. 290--306. Springer, 2020.

\bibitem{Koopman2022HowSafe}
P.~Koopman.
\newblock {\em How safe is safe enough}.
\newblock 2022.

\bibitem{Koopman2016}
P.~Koopman and M.~Wagner.
\newblock Challenges in autonomous vehicle testing and validation.
\newblock {\em SAE International Journal of Transportation Safety},
  4(1):15--24, 2016.

\bibitem{Koopman2020c}
P.~Koopman and M.~D. Wagner.
\newblock Positive trust balance for self-driving car deployment.
\newblock In {\em SAFECOMP Workshops}, vol. 12235 of {\em Lecture Notes in
  Computer Science}, pp. 351--357. Springer, 2020.

\bibitem{korssen2017systematic}
T.~Korssen, V.~Dolk, J.~van~de Mortel-Fronczak, M.~Reniers, and M.~Heemels.
\newblock Systematic model-based design and implementation of supervisors for
  advanced driver assistance systems.
\newblock {\em IEEE Transactions on Intelligent Transportation Systems},
  19(2):533--544, 2017.

\bibitem{Mehmed2019}
A.~Mehmed, W.~Steiner, M.~Antlanger, and S.~Punnekkat.
\newblock System architecture and application-specific verification method for
  fault-tolerant automated driving systems.
\newblock In {\em Intelligent Vehicles Symposium}, pp. 39--44. IEEE, 2019.

\bibitem{Mehmed2020}
A.~Mehmed, W.~Steiner, and A.~{\v{C}}au{\v{s}}evi{\'c}.
\newblock Systematic false positive mitigation in safe automated driving
  systems.
\newblock In {\em International Symposium on Industrial Electronics and
  Applications}, pp. 1--8. IEEE, 2020.

\bibitem{MobilEye2022TrueRedundancy}
Mobileye.
\newblock True redundancy, 2018.

\bibitem{Nister2019}
D.~Nistér, H.~Lee, J.~Ng, and Y.~Wang.
\newblock The safety force field, 2019.

\bibitem{nitsche2017pre}
P.~Nitsche, P.~Thomas, R.~Stuetz, and R.~Welsh.
\newblock Pre-crash scenarios at road junctions: A clustering method for car
  crash data.
\newblock {\em Accident Analysis \& Prevention}, 107:137--151, 2017.

\bibitem{oliveira2023architectures}
R.~G.~D. Oliveira, N.~Navet, and A.~Henkel.
\newblock Multi-objective optimization for safety-related available e/e
  architectures scoping highly automated driving vehicles.
\newblock {\em ACM transactions on Design Automation of Electronic Systems},
  2023.

\bibitem{osborne2022analysing}
M.~Osborne, R.~Hawkins, and J.~McDermid.
\newblock Analysing the safety of decision-making in autonomous systems.
\newblock In {\em International Conference on Computer Safety, Reliability, and
  Security}, pp. 3--16. Springer, 2022.

\bibitem{pek2020fail}
C.~Pek and M.~Althoff.
\newblock Fail-safe motion planning for online verification of autonomous
  vehicles using convex optimization.
\newblock {\em IEEE Transactions on Robotics}, 37(3):798--814, 2020.

\bibitem{pek2020FailSafeTrajectory}
C.~Pek, S.~Manzinger, M.~Koschi, and M.~Althoff.
\newblock Using online verification to prevent autonomous vehicles from causing
  accidents.
\newblock {\em Nature Machine Intelligence}, 2(9):518--528, 2020.

\bibitem{qiu2021reliability}
M.~Qiu, P.~Bazan, T.~Antesberger, F.~Bock, and R.~German.
\newblock Reliability assessment of multi-sensor perception system in automated
  driving functions.
\newblock In {\em 2021 IEEE 26th Pacific Rim International Symposium on
  Dependable Computing}, pp. 104--112. IEEE, 2021.

\bibitem{radlak2020organization}
K.~Radlak, M.~Szczepankiewicz, T.~Jones, and P.~Serwa.
\newblock Organization of machine learning based product development as per iso
  26262 and iso/pas 21448.
\newblock In {\em 2020 IEEE 25th Pacific Rim International Symposium on
  Dependable Computing}, pp. 110--119. IEEE, 2020.

\bibitem{Richards2010}
D.~Richards.
\newblock Relationship between speed and risk of fatal injury: pedestrians and
  car occupants.
\newblock 2010.

\bibitem{saberi2020beyond}
A.~K. Saberi, J.~Hegge, T.~Fruehling, and J.~F. Groote.
\newblock Beyond sotif: Black swans and formal methods.
\newblock In {\em 2020 IEEE International Systems Conference}, pp. 1--5. IEEE,
  2020.

\bibitem{SAEJ3016_2021}
SAE.
\newblock Taxonomy and definitions for terms related to driving automation
  systems for on-road motor vehicles.
\newblock 2021.

\bibitem{salay2022safety}
R.~Salay and K.~Czarnecki.
\newblock A safety assurable human-inspired perception architecture.
\newblock {\em arXiv preprint arXiv:2205.07862}, 2022.

\bibitem{schuster2022formally}
T.~Schuster, E.~Seferis, S.~Burton, and C.-H. Cheng.
\newblock Formally compensating performance limitations for imprecise 2d object
  detection.
\newblock In {\em International Conference on Computer Safety, Reliability, and
  Security}, pp. 269--283. Springer, 2022.

\bibitem{Shalev-Shwartz2017}
S.~Shalev-Shwartz, S.~Shammah, and A.~Shashua.
\newblock On a formal model of safe and scalable self-driving cars, 2018.

\bibitem{tamke2011flexible}
A.~Tamke, T.~Dang, and G.~Breuel.
\newblock A flexible method for criticality assessment in driver assistance
  systems.
\newblock In {\em 2011 IEEE Intelligent Vehicles Symposium}, pp. 697--702.
  IEEE, 2011.

\bibitem{Matlab2021}
{The Mathworks, Inc.}
\newblock Matlab and automated driving toobox release 2021b.
\newblock 2021.

\bibitem{Torngren2019}
M.~T{\"o}rngren, X.~Zhang, N.~Mohan, M.~Becker, L.~Svensson, X.~Tao, D.~Chen,
  and J.~Westman.
\newblock Architecting safety supervisors for high levels of automated driving.
\newblock In {\em 21st International Conference on Intelligent Transportation
  Systems}, pp. 1721--1728. IEEE, 2018.

\bibitem{ul4600}
{UL4600, Standard for Evaluation of Autonomous Products}, {2020}.

\bibitem{Horst1990PET}
A.~Van~der Horst.
\newblock A time-based analysis of road user behaviour in normal and critical
  encounters.
\newblock {\em Delft, The Netherlands}, p.~78, 1990.

\bibitem{Wang2019}
Y.~Wang, Z.~Liu, Z.~Zuo, Z.~Li, L.~Wang, and X.~Luo.
\newblock Trajectory planning and safety assessment of autonomous vehicles
  based on motion prediction and model predictive control.
\newblock {\em IEEE Transactions on Vehicular Technology}, 68(9):8546--8556,
  2019.

\bibitem{WaymoSafety2021}
Waymo.
\newblock Safety report 08-2021.
\newblock techreport, Waymo, Aug. 2021.

\bibitem{weast2020sensors}
J.~Weast.
\newblock Sensors, safety models and a system-level approach to safe and
  scalable automated vehicles.
\newblock {\em arXiv preprint arXiv:2009.03301}, 2020.

\bibitem{WHO2018}
{World Health Organisation}.
\newblock Global status report on road safety.
\newblock 2018.

\bibitem{wozniak2020safety}
E.~Wozniak, C.~C{\^a}rlan, E.~Acar-Celik, and H.~J. Putzer.
\newblock A safety case pattern for systems with machine learning components.
\newblock In {\em International Conference on Computer Safety, Reliability, and
  Security}, pp. 370--382. Springer, 2020.

\bibitem{zeller2022component}
M.~Zeller.
\newblock Component fault and deficiency tree (cfdt): Combining functional
  safety and sotif analysis.
\newblock In {\em International Symposium on Model-Based Safety and
  Assessment}, pp. 146--152. Springer, 2022.

\end{thebibliography}

\appendix
\section{Risk computation method}
Below we will address the risk computation method as used in this publication. 
\subsection{Probability computations}
\label{appendix:risk}
Here we will show how we create usable translations from conventional safety indicator values to a relative probability value, as used for \eqref{eq:risk_simplified}. 
Prior, we introduced the indices E as an index of the relevant event, $i$ as the index of the channel of which the WM$_i$ is used to evaluate the risk and $j$ as the index of the channel of which the trajectory $T_j$ is evaluated.
To make the following equations easier to read, we introduce the dummy index $\xi$ to replace the combination of indices $\mathrm{E},ij$.
Following e.g., \cite{Damerow2015risk,Eggert2014risk}, we can create a map to transform conventional safety indicator values into an approximate relative probability, via
\begin{equation}
    \mathcal{F}_\gamma: x_{\xi\gamma}(\tau) \rightarrow P_{\xi\gamma}(\tau)    
\end{equation}
where $x_{\xi\gamma}(\tau)\Leftrightarrow x_{\mathrm{E},ij\gamma}(\tau)$ is a dummy variable that represents the value of a conventional safety indicator $\gamma \in \{1,2,...,n_{\mathrm{si}} \}$, with $n_{\mathrm{si}}$ the number of used safety indicators, at predicted time $\tau$, according to WM$_i$ and tested trajectory $T_j$ 
and $P_{\xi\gamma}(\tau) \Leftrightarrow P_{\mathrm{E},ij\gamma}(\tau)$ is the resultant probability of an adverse event $\mathrm{E}$ to occur at the indicated predicted time interval.
The mapping operator $\mathcal{F}_\gamma$ is implemented through the logistic function, parameterized as
\begin{equation}
    P_{\xi\gamma}(\tau) = \frac{1/\Delta_{\mathrm{p}}}{1 + e^{-\beta_\gamma \left(x_{\xi\gamma}(\tau)-x_{\gamma ,0} \right)}}
    \label{eq:singleSIProbability}
\end{equation}
where $\Delta_{\mathrm{p}}$ is the prediction timestep that governs the length of the relevant interval and $\beta_\gamma $ and $x_{\gamma ,0}$ represent tuning factors specific for safety indicator $\gamma $.
The various used safety indicators and their mapped probabilities as calculated through \eqref{eq:singleSIProbability} are combined to a single probability value for event E through
\begin{equation}
    P_{\xi}(\tau) =
        \argmin \left(1, \sum^{n_{\mathrm{si}}}_{\gamma =1}
        P_{\xi\gamma}(\tau)
        \right)
        \Phi_{\xi}(k)
    \label{eq:probabilityEq}
\end{equation}
{where $\Phi_{\xi}(k)$ represents the probability of existence of an object, scaling the overall probability of collision linearly.}

By selecting safety indicators $\gamma \in \{1,2,...,n_{\mathrm{si}} \}$ whose applicable domains do not overlap for safe to moderately dangerous trajectories,  \eqref{eq:probabilityEq} will be consistent. 
We refer to this selection of safety indicators as effectively decoupled.
\begin{figure}[ht]
    \centering
    \includegraphics[width=6.5cm]{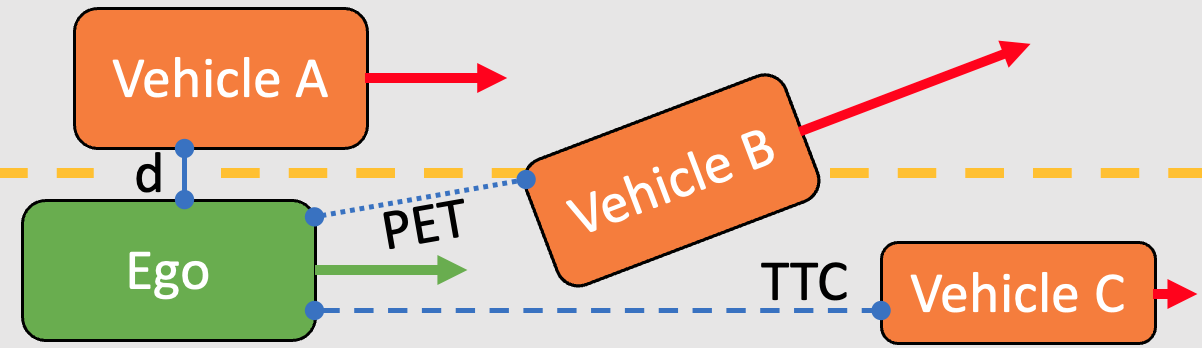}
    \caption{The approximately decoupled set of safety indicators.}
    \label{fig:D_PET_TTC_Explained}
\end{figure}  
Fig. \ref{fig:D_PET_TTC_Explained} shows an example of a group of safety indicators that can be tuned to be effectively decoupled, with the distance (d, indicated to \textit{Vehicle A}) tuned to be relevant for passing objects at close lateral distances, but, through its tuning, irrelevant for objects on path when traveling at speeds {beyond 1 m/s}, 
the Time-To-Collision indicator\cite{tamke2011flexible} (TTC, indicated to \textit{Vehicle C}) for driving up to objects on the ego path that have a lower speed than the ego vehicle, but undefined for vehicles with higher speeds (\textit{Vehicle B}) or vehicles tangent to the ego path (\textit{Vehicle A}) and 
the Post-Encroachment-Time indicator \cite{Horst1990PET}, an indicator of the amount of time passed between an object occupying a space and the ego vehicle occupying at least some part of that same space (PET, indicated to \textit{Vehicle B}), relevant for any objects whose path is crossed irrespective of the objects relative speed to the ego vehicle, but irrelevant to vehicles tangent to the ego's path (\textit{Vehicle A}) and not sensitive enough for slower vehicles in front (\textit{Vehicle C}). 
In case a trajectory evaluated through \eqref{eq:probabilityEq} becomes very dangerous at some future predicted time interval starting at $\tau$, \eqref{eq:probabilityEq} saturates to unity at $\tau$.
In those cases, the overlap between the safety indicators is not relevant, as a (near-)certain collision is predicted.
Parameters in \eqref{eq:probabilityEq} are manually tuned, to attain what is judged as reasonable probability approximations. 
Though the examples shown in Fig. \ref{fig:D_PET_TTC_Explained} are indicated with respect to vehicles, the safety indicators can also be used with respect to other types of objects, road boundaries or other elements.
The safety indicator parameters used in \eqref{eq:probabilityEq} are shown in Table \ref{tab:probOfCollisionParams}.
\begin{table}[]
\centering
\caption{Probability of collision parameters as used.}
\label{tab:probOfCollisionParams}
\begin{tabular}{@{}lll@{}}
\toprule
Safety Indicator & $\beta_\gamma$ & $x_{_\gamma,0}$ \\ \midrule
TTC & 4 & 2.5 \\
PET & 20 & 0.3 \\
distance & 11 & 0.5 \\ \bottomrule
\end{tabular}
\end{table}
\subsection{Severity score calculation}
\label{appendix:severityScore}
The severity $S_{\mathrm{E},ij}(\tau)$ for event E required for the overall risk is computed through a logistic function,
\begin{equation}
    S_{\mathrm{E},ij}(\tau) = \lambda_0\left( 1- \frac{\lambda_1}{1 + e^{-\lambda_2 \left(\Delta v(\tau)-\Delta v_{0} \right)}} \right)
    \label{eq:severityCalc}
\end{equation}
with $\lambda_0$, $\lambda_1$ and $\lambda_2$ and $\Delta v_{0}$ tuning factors dependent on the type of event or object that $S_{\mathrm{E},ij}$ refers to, e.g., vulnerable road user, vehicle, rule-restriction-boundary, while $\Delta v(\tau)$ represents the closing speed as a substitute for the expected impact speed.
The closing speed is a function of the change of distance $d$ between the closest boundaries of the two objects over time, predicted by WM$_i$ and $T_j$ and calculated via
\begin{equation}
    \Delta v(\tau) = \frac{\delta d(\tau)}{\delta \tau}
\end{equation}

\end{document}